\documentclass[sigconf]{acmart}
\usepackage[most]{tcolorbox}
\usepackage{pgfplots}
\usepackage{listings}
\usepackage{tabularx}
\usepackage{multirow}
\usepackage{subcaption}
\usepackage{booktabs}

\AtBeginDocument{%
  }

\renewcommand\footnotetextcopyrightpermission[1]{} 
\acmConference[KDD '26]{Proceedings of the 32nd ACM SIGKDD Conference on
Knowledge Discovery and Data Mining}{August 2026}{Location}
\acmYear{2026}
\copyrightyear{2027}

\begin{document}

\title{EarlyDx: An Admission-Anchored Benchmark for Open-Ended Generation of Evidence-Supported ED-Encounter Diagnoses}

\author{Jiahui Li}
\orcid{0009-0007-5129-0107}
\affiliation{%
  \institution{University of Georgia}
  \city{Athens}
  \country{USA}
}\email{jl57095@uga.edu}

\author{Ruili Fang}
\affiliation{%
  \institution{University of Georgia}
  \city{Athens}
  \country{USA}}
\email{ruili.fang@uga.edu}

\author{Zishuai Liu}
\affiliation{%
  \institution{University of Georgia}
  \city{Athens}
  \country{USA}
}\email{zishuai.liu@uga.edu}

\author{Yutong Guo}
\affiliation{%
  \institution{University of Georgia}
  \city{Athens}
  \country{USA}
}\email{yutong.guo@uga.edu}

\author{Nan Yang}
\affiliation{%
  \institution{Beijing Luhe Hospital}
  \city{Beijing}
  \country{China}
}\email{yangnan@mail.ccmu.edu.cn}

\author{Wenzhan Song}
\affiliation{%
  \institution{University of Georgia}
  \city{Athens}
  \country{USA}}
\email{wsong@uga.edu}

\author{Jin Lu}
\affiliation{%
  \institution{University of Georgia}
  \city{Athens}
  \country{USA}}
\email{jin.lu@uga.edu}

\author{Fei Dou}
\correspondingauthor
\affiliation{%
  \institution{University of Georgia}
  \city{Athens}
  \country{USA}}
\email{fei.dou@uga.edu}

\newcommand{\std}[1]{{\fontsize{7}{8.2}\selectfont$\pm$ #1}}
\newcommand{\val}[2]{#1 \std{#2}}
\newcommand{\valsig}[3]{#1\textsuperscript{#3} \std{#2}}
\renewcommand{\shortauthors}{Li et al.}

\begin{abstract}

Clinical diagnosis at hospital admission must be made rapidly from limited,
incomplete evidence. Existing diagnosis-prediction benchmarks are poorly suited
to this setting: they restrict prediction to closed code sets, exclude free-text
notes, and supervise with discharge diagnoses that incorporate the full inpatient
course. We introduce \textbf{EarlyDx}, a large-scale benchmark for
\emph{open-ended} early diagnosis, built from $154{,}834$ emergency department
encounters in MIMIC-IV. Each encounter is restricted to records available at admission time $t_0$ and supervised by the diagnoses recorded during the ED encounter
rather than at discharge. An LLM auditor further verifies every free-text label
as \emph{supported}, \emph{partially supported}, or \emph{unsupported} by that
evidence; the primary evaluation scores only fully supported labels.
Under a semantic LLM-as-judge protocol, no evaluated system---frontier general,
medical-specialized, or in-domain post-trained---synthesizes admission-time
evidence reliably. Zero-shot models score largely by extraction, recovering only
$3$--$31\%$ of diagnoses that must be inferred rather than read from the record;
post-training raise inference-dependent recall to $56\%$, but a sizeable margin
remains, and on time-critical conditions no system attains a clinician's balance
of sensitivity and precision. 
We release the full construction and evaluation
pipeline at \href{https://github.com/jimmylihui/EarlyDx}{here}.

\end{abstract}




\maketitle

\newcommand{\cmark}{\ding{51}}\newcommand{\xmark}{\ding{55}}

\begin{table*}[t]\centering\small
\caption{EarlyDx versus existing medical-LLM benchmarks.
\emph{Task}: Gen.\ = generation of free-text diagnoses, Cls.\ = classification.
\emph{Label time}: when the label is set (early/admission vs.\ discharge).
``Multi-source text'': labs, imaging/ECG interpretations, vitals, and history
rendered as text. \emph{Evid.\ verif.}: evidence-grounded label verification.}
\label{tab:benchmarks}
\setlength{\tabcolsep}{4pt}
\begin{tabular}{l ccccccc c c}
\toprule
& Med-PaLM 2 & Med-PaLM M & Wang et al. & AgentClinic & MIRA & MedTVT & MDS-ED & MC-BEC & \textbf{EarlyDx}\\
& \cite{singhal2025toward} & \cite{tu2024towards} & \cite{wang2025novel} & \cite{schmidgall2024agentclinic} & \cite{ferber2026towards} & \cite{zhang2026medtvt} & \cite{alcaraz2025enhancing} & \cite{chen2023multimodal} & \textbf{(ours)}\\
\midrule
Task            & QA & Mixed & QA & Agent & Agent & Cls. & Cls. & Cls. & \textbf{Gen.}\\
CoT             & \xmark & \xmark & \xmark & \xmark & \xmark & \xmark & \xmark & \xmark & \cmark\\
Labels          & Closed & Closed & Open & Closed & Closed & Closed & Closed & Closed & \textbf{Open}\\
Evid.\ verif.   & \xmark & \xmark & \xmark & \xmark & \xmark & \xmark & \xmark & \xmark & \textbf{\cmark}\\
Modalities      & Text & Text+Img+Omics & Text & Text+Img & Text/FHIR & ECG+CXR+Lab & Tab+ECG & Multimodal & \textbf{Multi-src text}\\
Label time      & -- & -- & -- & Interact. & Discharge & Discharge & Discharge & Discharge & \textbf{Early}\\
Scale           & $\sim$10k Q & 14 tasks & -- & 534 & 574 & 8.7k & 129k & 103k & \textbf{155k}\\
\bottomrule
\end{tabular}
\end{table*}

\begin{figure}[t]
  \centering
  \begin{subfigure}[b]{0.23\textwidth}
    \centering
    \includegraphics[width=\linewidth]{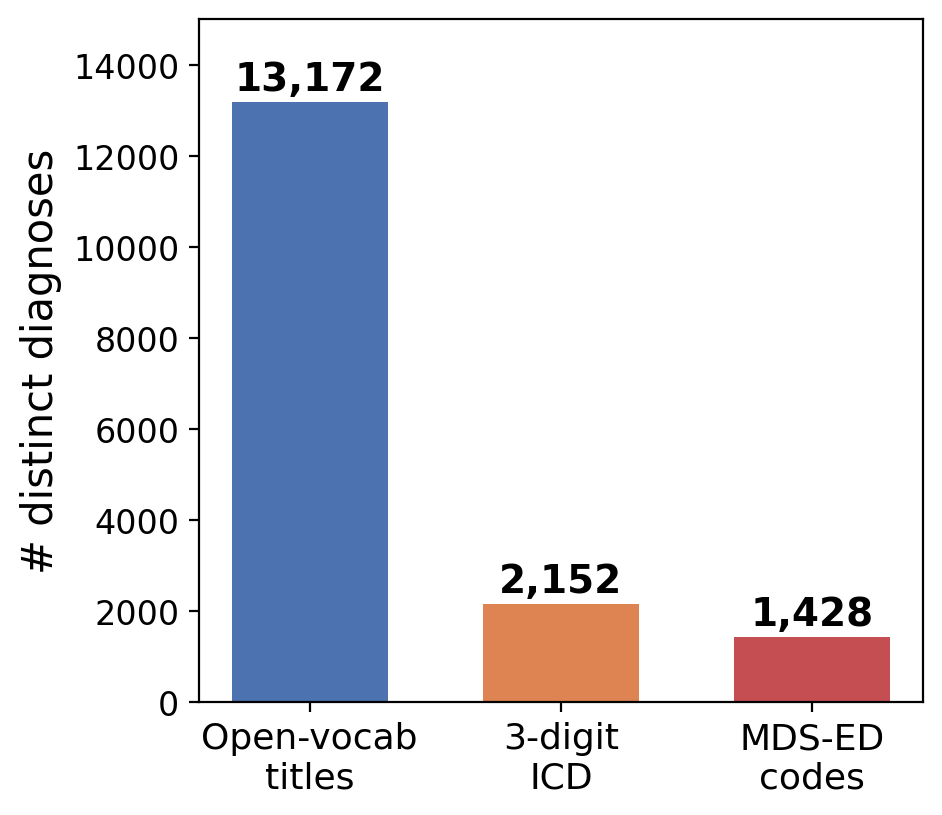}
    \caption{Granularity lost to closed-set ICD conversion.}
    \label{fig:icd_loss_bar}
  \end{subfigure}
  \hfill
  \begin{subfigure}{0.23\textwidth}
  \centering
  \includegraphics[height=3.5cm]{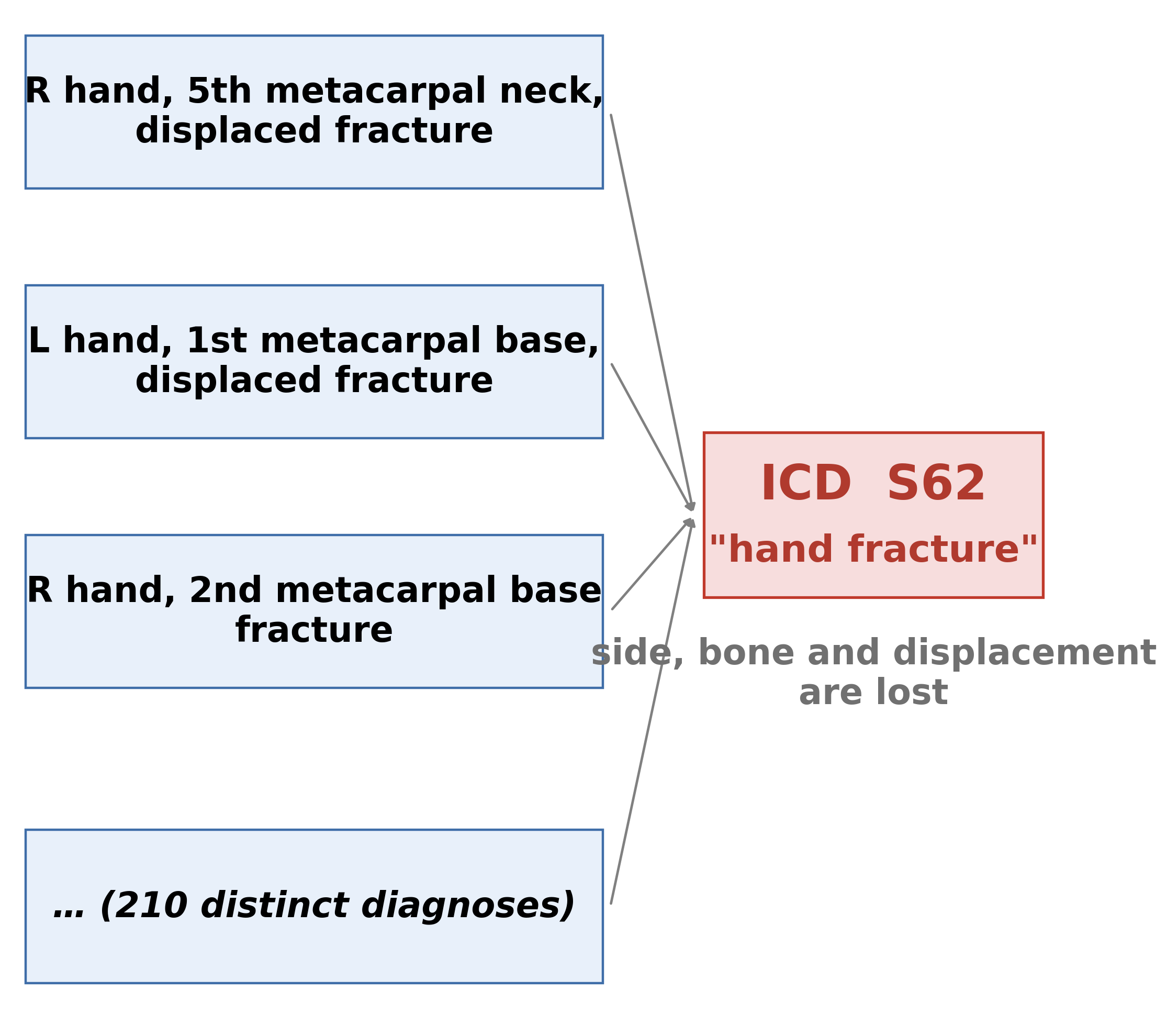}
  \caption{One category (S62) absorbs 210 distinct diagnoses.}
    \end{subfigure}
  \caption{Information lost when free-text diagnoses are converted to a closed ICD
  code set. (a) The $13{,}172$ distinct ED diagnosis titles collapse to $2{,}152$
  three-digit categories ($6.1\times$) and to MDS-ED's $1{,}428$ codes
  ($9.2\times$). (b) S62 conflates $210$ clinically distinct diagnoses, discarding
  laterality, bone, and displacement that open-ended free-text labels preserve.}
  \label{fig:icd_loss}
\end{figure}

\section{Introduction}
When a patient is admitted through the emergency department, clinicians must
commit to an early, working diagnosis under uncertainty: one formed from only the
evidence available at the point of admission---vitals, laboratory results,
imaging, and prior history---to justify admission and guide the immediate
``next action'' in care~\cite{zikos2019estimation,nagar2026clr,stinard2026clinicalbench}.
This early assessment differs fundamentally from a final or discharge diagnosis,
which is established retrospectively after the full hospital course; it is instead
a synthesis of partially observed signals under substantial
clinical uncertainty~\cite{nagar2026clr}. 

Large language models (LLMs) have shown strong capabilities across diverse medical tasks~\cite{lee2024llm,van2023radadapt,tu2024towards} and are natural candidates for this synthesis-under-uncertainty problem. However, existing benchmarks for early diagnosis are designed for non-LLM, closed-set classifiers~\cite{chen2023multimodal,alcaraz2024mds,sundrani2023predicting} and suffer from three limitations that make them ill-suited for evaluating modern LLMs. \textbf{(i) Closed, converted labels:} diagnoses are mapped to a fixed ICD code set, discarding clinical nuance and information lost during text-to-code conversion~\cite{o2005measuring}, as shown in Figure~\ref{fig:icd_loss}. \textbf{(ii) Restricted, non-textual modalities:} rich free-text signals such as chief complaints, radiology findings, and prior history are excluded in favor of tabular features. \textbf{(iii) Temporal inconsistency:} models receive early, admission-time
inputs but are supervised with \emph{discharge} diagnoses that are coded over the
\emph{entire} hospital course. These labels depend on information unavailable at
admission.

To address these gaps, we introduce \textbf{EarlyDx}, a large-scale
($N{=}154{,}834$ encounters), admission-anchored benchmark for \emph{open-ended}
early diagnosis. Its central design choice is that recorded diagnoses are not
taken as ground truth uncritically: an LLM auditor verifies each free-text label
as \emph{supported}, \emph{partially supported}, or \emph{unsupported} by the
admission-time evidence, so the benchmark can score models only on diagnoses the
early record actually substantiates. This evidence grounding is what makes the
benchmark diagnostic rather than merely predictive---it lets us separate
diagnoses a model \emph{extracts} from text from those it must \emph{infer}.

\noindent Our contributions are:
\begin{itemize}
  \item \textbf{An admission-anchored benchmark for open-ended early diagnosis.}
  154,834 ED encounters supervised with \emph{ED-encounter} rather than discharge
  diagnoses, all evidence clipped to admission ($W{=}0$), labels kept as free
  text, scored by a semantic judge (89.7\% agreement with a second judge, 94\%
  with manual matching).
  \item \textbf{Labels grounded in the evidence available at admission.} Each
  label is verified against the early record; the primary track scores the 32.7\%
  fully supported, the 37.0\% partially supported form a secondary track. Systems
  separate sharply across this boundary (0.51 vs.\ 0.27 F1), confirming it is
  meaningful.
  \item \textbf{Current LLMs extract diagnoses rather than infer them.} Only 43\%
  of supported diagnoses appear verbatim in the input; zero-shot LLMs recover
  37--65\% of these but 3--31\% of the implicit majority. Post-training lifts
  implicit recall to 56\% without closing the gap, and no system, ours included,
  reaches a clinician's operating point on time-critical conditions.
\end{itemize}

\section{Related Work}
\subsection{LLMs for Diagnosis}
Large language models achieve expert-level accuracy on medical question
answering~\cite{singhal2025toward}, but QA is typically multiple-choice or
short-answer and does not capture the evidence synthesis and uncertainty of real
practice. Later work moves to open-ended clinical scenarios, where domain-specific
medical LLMs show consistent advantages over general-purpose
models~\cite{wang2025novel}, and to autonomous agents that navigate a clinical
action space to reach a diagnosis~\cite{ferber2026towards,schmidgall2024agentclinic,tu2024towards}.
Because the underlying datasets contain no doctor--patient dialogue, these agentic
settings rely on rule-based or LLM-driven patient simulators to synthesize the
missing interaction, yielding trajectories that depart from authentic clinical
documentation. We instead evaluate models directly on real admission-time records,
isolating diagnostic capability from the artifacts of simulated interaction.

\subsection{Early Diagnosis Benchmarks}
Existing emergency-medicine benchmarks observe the patient over a short window
measured from ED \emph{arrival}: MC-BEC predicts decompensation, disposition, and
revisit from the first 15 minutes after rooming~\cite{chen2023multimodal}, and
MDS-ED predicts deterioration and a fixed set of $1{,}428$ ICD-10-CM discharge
diagnoses from the first 90 minutes~\cite{alcaraz2025enhancing}. Such windows
precede much of the diagnostic workup---labs and imaging are typically not yet
back---while the supervision comes from discharge codes assigned retrospectively
over the whole inpatient stay, and therefore includes conditions unknowable at
admission.
EarlyDx differs on all three axes (Table~\ref{tab:benchmarks}): it anchors at the
\emph{admission decision} ($t_0{=}\texttt{admittime}$), where the complete ED
workup is observable but no post-admission record is; it supervises with
\emph{ED-encounter} diagnoses as open-ended text rather than closed-set codes;
and it is the only benchmark to verify each label against the evidence actually
available at $t_0$.
\section{Method}

\begin{figure*}[t]
\vspace{-3pt}
    \centering
    \includegraphics[width=0.89\textwidth]{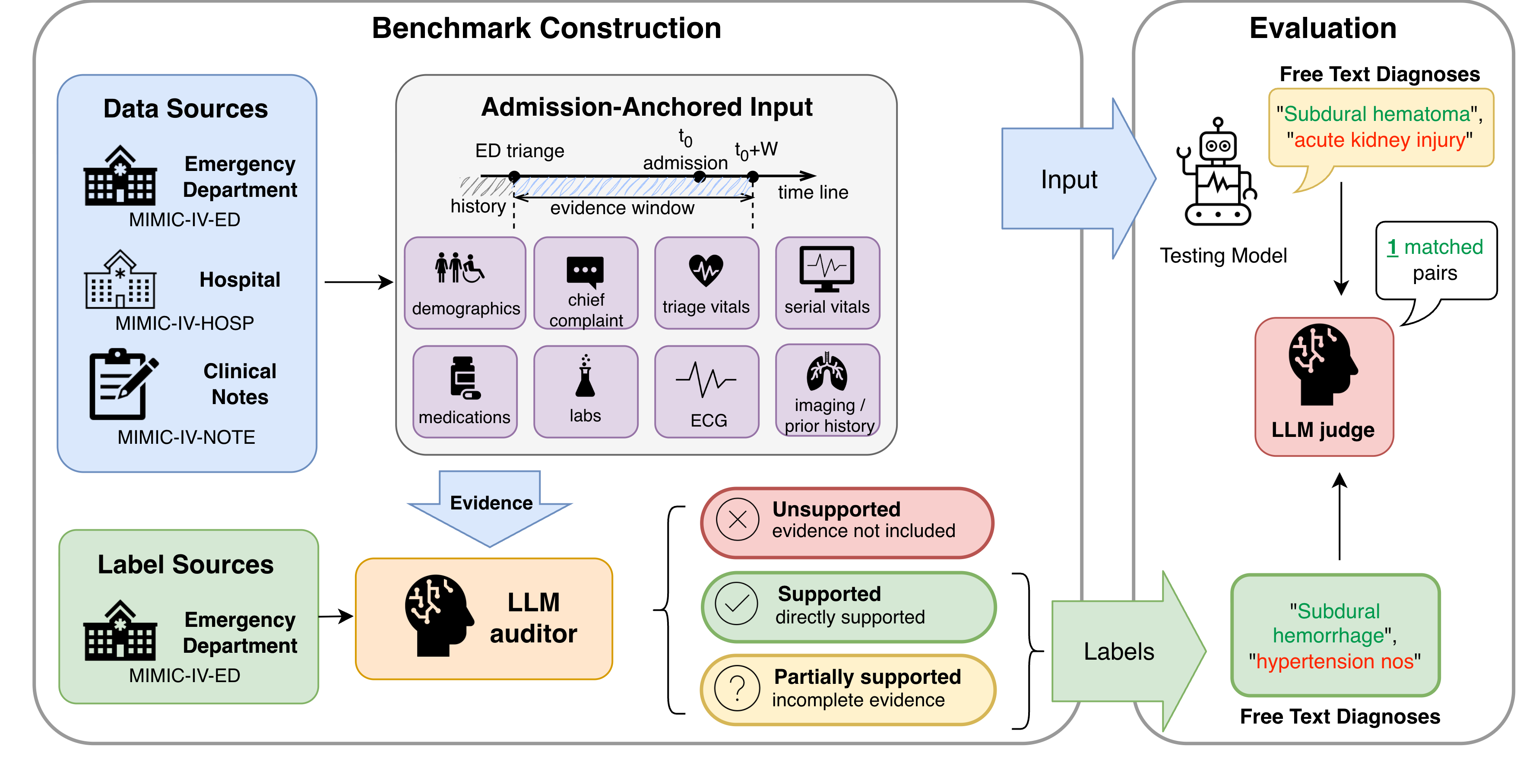}
    \caption{Overview of the EarlyDx framework. From MIMIC-IV emergency-department
encounters, we build an admission-anchored early-workup input (cutoff at
$t_0{+}W$; the main benchmark uses $W{=}0$, i.e., the cutoff coincides with the
admission time $t_0$), verify free-text diagnosis labels into evidence-grounded
support categories, generate gold-conditioned chain-of-thought supervision for
supported and partially supported cases, split the evidence-supported set, and evaluate zero-shot and post-trained LLMs via semantic
LLM-as-judge matching.}
    \label{fig:earlydx_framework}
\end{figure*}

\subsection{Cohort and Task Formulation}
Figure~\ref{fig:earlydx_framework} summarizes the construction and evaluation pipeline. EarlyDx is derived from MIMIC-IV~\cite{johnson2023mimic}, combining the ED,
hospital, and note modules. We include every emergency-department stay that
results in a hospital admission, yielding $N{=}154{,}834$ encounters. We formulate
early diagnosis as an \emph{open-ended, multi-label} task: given the information
available at admission $x$, a model produces a set of free-text diagnoses
$\hat{Y}{=}\{\hat{y}_1,\dots,\hat{y}_k\}$, evaluated against the reference set $Y$
of \emph{ED-encounter} documented diagnoses.

\subsection{Admission-Anchored Early Workup Input Construction}
\begin{figure}[ht]\centering
  \includegraphics[width=0.95\columnwidth]{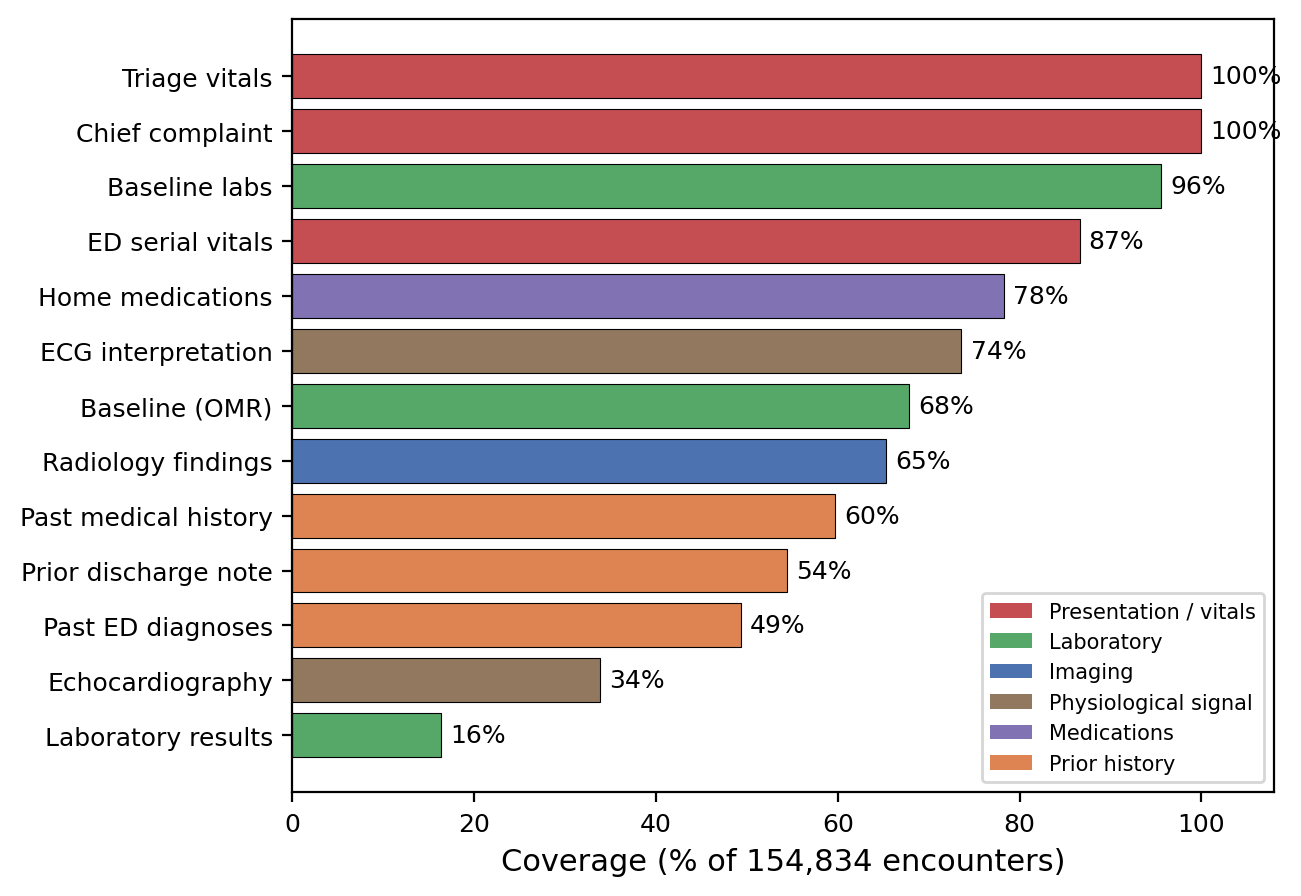}
  \caption{Modality coverage in EarlyDx at $W{=}0$: fraction of the $154{,}834$
  encounters for which each input field is present. Presentation and triage vitals
  are universal; ECG and radiology cover $74\%$ and $65\%$; in-window laboratory
  results only $16\%$, since admitted-patient results are timestamped at or after
  $t_0$. Colors group modalities by type.}
  \label{fig:coverage}
  \Description{Horizontal bar chart of input-field coverage in the EarlyDx cohort,
  colored by modality type.}
\vspace{-8pt}
\end{figure}

We set \(t_0\) to the hospital admission time (\texttt{admittime}) and retain only
records charted at or before \(t_0{+}W\). Each encounter is serialized into a
single text prompt comprising: (i)~demographics and arrival mode; (ii)~chief
complaint and triage vitals; (iii)~ED serial vitals; (iv)~home medications
(\texttt{medrecon}); (v)~baseline measurements (\texttt{OMR}); (vi)~laboratory
results; (vii)~ECG interpretation (\texttt{machine\_measurements});
(viii)~echocardiography; (ix)~radiology findings; and (x)~prior history (past ED
diagnoses and prior discharge problem lists). Imaging and physiological signals
are rendered as report findings and structured measurements rather than raw pixels
or waveforms; fields with no qualifying record are marked \texttt{None}. Appendix~\ref{app:example} shows a complete serialized encounter. Coverage
is uneven (Figure~\ref{fig:coverage}): chief complaint and triage vitals are
universal, ECG and radiology reach $74\%$ and $65\%$, echocardiography $34\%$, and
in-window laboratory results only $16\%$---admitted-patient results are typically
timestamped at or after \(t_0\). This heterogeneity is intrinsic to real admission
data and is what makes the task one of reasoning over partially observed evidence.

\noindent\textbf{Temporal cutoff.} Because the ED diagnosis table records no
timestamp for when a diagnosis was formulated, $t_0$ is an \emph{observable} anchor
for the early diagnostic context rather than the true decision time; any window
beyond it may therefore expose post-decision evidence. We treat $W$ as a parameter
and study its effect directly (Table~\ref{tab:window}). The main benchmark uses the
most conservative setting, $W{=}0$. Since $t_0$ falls at the end of the ED workup
(median ED stay ${\approx}4.6$\,h), this still admits the full pre-admission
evaluation, minimizing known leakage without leaving the input evidence-poor;
wider windows ($W\in\{6,24\}$\,h) are reported in
Appendix~\ref{app:evidence_window}.

\noindent\textbf{Timestamp semantics.} Evidence is filtered on event time:
\texttt{charttime} for laboratory results, vital signs, and radiology reports;
\texttt{ecg\_time} for ECG interpretations; and \texttt{chartdate} for outpatient
measurements. Event time precedes the time at which a result becomes readable in
the record (\texttt{storetime}) by a median of $1.2$\,h for laboratory results and
$1.8$\,h for radiology reports, so event-time filtering is the more inclusive of
the two conventions and our figures should be read as an upper bound on what was
strictly available at $t_0$; Appendix~\ref{app:store} reports a sensitivity
analysis under \texttt{storetime} filtering. Prior history is drawn only from earlier admissions whose \texttt{admittime} precedes $t_0$.

\subsection{Diagnosis Labels and Evidence-Grounded Verification}
Reference diagnoses are the ICD~\cite{hirsch2016icd} titles from the MIMIC-IV-ED
\texttt{diagnosis} table for the corresponding stay, taken as open-vocabulary
text. This table records the diagnoses \emph{billed for the emergency encounter}
(ordered by \texttt{seq\_num}, without a per-diagnosis timestamp) and is therefore
scoped to the early presentation rather than the full hospitalization. We filter
administrative and symptom codes (ICD-10 chapters R, V--Z; ICD-9 E/V, 780--799)
and over-generic Not Otherwise Specified (NOS) / Not Elsewhere Classified (NEC)
entries, retaining genuine diagnoses.
Because these labels are coded administratively and retrospectively, some cannot
be inferred at admission. We therefore apply an LLM auditor
(MiniMax~M3~\cite{minimax2025minimaxm1scalingtesttimecompute}) that, given the
admission-time input $x$ and a reference label $y$, assigns one of three
categories. A label is \textbf{supported} when a specific admission-time
finding---in the labs, imaging report, ECG interpretation, or vitals---directly
substantiates it. It is \textbf{partially supported} when the evidence is only
indirect or circumstantial (a relevant comorbidity in the past history, a
corresponding home medication, a suggestive but non-confirmatory sign), so that no
admission-time finding establishes it definitively. It is \textbf{unsupported}
when it cannot be derived from $x$ at all, because it depends on information
unavailable at admission: results arriving later in the stay such as microbiology
cultures, the post-admission workup, or outside-hospital records (e.g.\ a transfer
patient whose confirmatory imaging was performed externally). These categories account for $32.7\%$, $37.0\%$, and $30.4\%$ of the $242{,}729$
labels. Appendix~\ref{app:boundary} reports an independent re-audit of this assignment by
four clinicians and shows that system rankings are invariant under a stricter,
human-confirmed gold standard; Appendix~\ref{app:review} gives the review protocol.
We define three evaluation tracks. The \textbf{primary} track scores against
supported labels only; the \textbf{secondary} track adds partially supported
labels; and a \textbf{historical-comorbidity} track isolates the partially
supported labels alone. Unsupported labels are excluded throughout, and headline
numbers use the primary track unless noted. Appendix~\ref{app:ignore} shows that rankings are unchanged if predictions matching
partially supported labels are withheld from the precision denominator rather than
counted as false positives.

\begin{figure}[t]\centering
  \includegraphics[width=0.73\columnwidth]{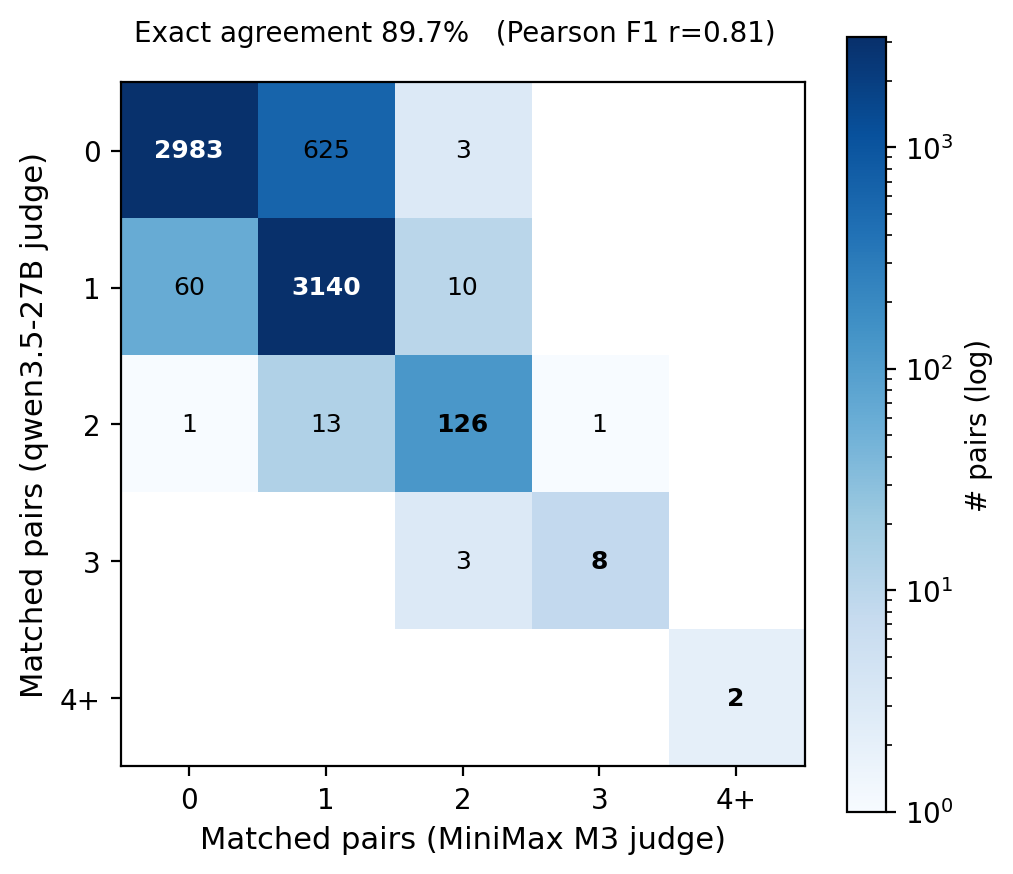}
  \caption{Agreement between two independent LLM judges on the matched-pair count over all
  6{,}975 test predictions.}
  \label{fig:judge}
\end{figure}

\begin{figure}[t]
  \centering
  \includegraphics[width=0.58\linewidth]{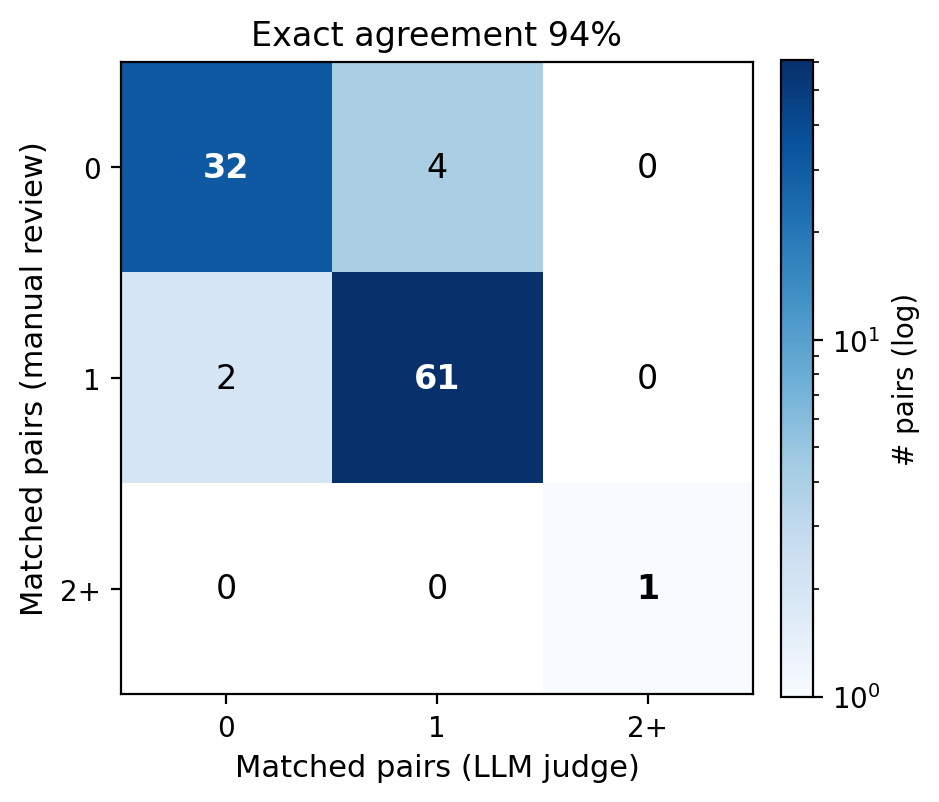}
  \caption{Agreement between the LLM judge and manual gold-matching review on
  100 test predictions. }
  \label{fig:judge_human}
\vspace{-5pt}
\end{figure}

\subsection{Evidence Availability Across Patient Subgroups}
\label{sec:subgroup}
Because the evidence class determines which labels enter the primary track, a
systematic difference in evidence availability across patient groups would bias
the benchmark itself. Appendix~\ref{app:subgroup} reports the class distribution by subgroup alongside
the cohort's composition. The cohort is adult and skewed toward
higher acuity---$57\%$ of encounters are triaged ESI~1--2 and $53\%$ arrive by
ambulance---as expected of ED visits ending in admission; race and language
distributions reflect the source institution's catchment and are not nationally
representative.

We find no difference of consequence. Supported rates vary by at most $5$
percentage points across any grouping and not at all by sex ($p{=}0.32$). What residual
variation there is follows clinical rather than demographic axes---ESI~1
presentations have the highest supported rate, consistent with a more extensive
workup---so evidence availability varies more with how long one observes than with
who the patient is. Model performance, by contrast, is not flat across subgroups: Appendix~\ref{app:subgroup}
reports a monotone decline with patient age that appears in every system, including
models never trained on EarlyDx.

\subsection{Evaluation Protocol}
\label{sec:eval-protocol}
Because diagnoses are free text, exact string matching is unreliable (e.g.\
``CAD'' vs.\ ``coronary artery disease''). We therefore adopt an LLM-as-judge
protocol: for each encounter the judge computes a one-to-one matching between
$\hat{Y}$ and $Y$ and returns the number of matched pairs $m$. We report micro-
and example-based Precision, Recall, and F1, with $P{=}m/|\hat{Y}|$ and
$R{=}m/|Y|$. The judge is fixed (qwen3.5-27B~\cite{team2026qwen3})
and its decisions are cached for reproducibility.

Two checks establish that the resulting scores are neither an artifact of a single
model nor merely self-consistent. Re-scoring all $6{,}975$ non-empty (gold,
prediction) pairs with a second, independent judge yields exact agreement on the
matched-pair count in $89.7\%$ of cases and agreement within one pair in $99.9\%$,
at a per-example F1 correlation of $r{=}0.81$ (Figure~\ref{fig:judge}). A manual
gold-matching review of $100$ sampled pairs matches the judge's count in $94\%$ of
cases, all within one pair (Figure~\ref{fig:judge_human}).

\section{Experiment}
\subsection{Dataset Split}
Discarding unsupported labels leaves $120{,}789$ of the $154{,}834$ admitted ED
encounters with at least one supported or partially supported label. We split these by patient---all encounters of a
subject fall entirely in train or test---giving $113{,}814$ training and $6{,}975$
test encounters.

\begin{figure}[t]
  \centering
  \includegraphics[width=\linewidth]{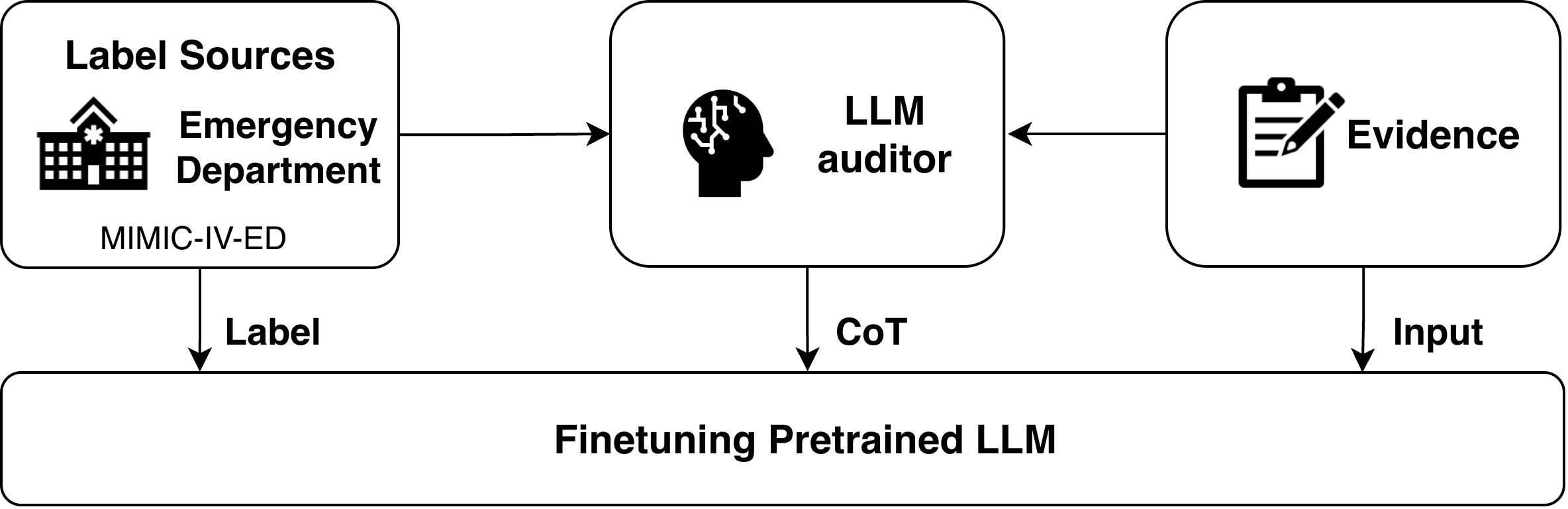}
  \caption{The EarlyDx fine-tuning framework. For each encounter, an ED-encounter
diagnosis label (MIMIC-IV-ED) and the admission-time evidence are passed to an
LLM auditor that produces a gold-conditioned chain-of-thought (CoT) rationale.
The evidence (input), label (target), and CoT (rationale) together fine-tune a
pretrained LLM to predict diagnoses with an inspectable rationale.}
  \label{fig:finetuning}
\end{figure}

\subsection{Models}
As \textbf{zero-shot} baselines we prompt state-of-the-art LLMs to produce
diagnoses directly, spanning general-purpose models---\textsc{GPT-5.5}~\cite{singh2025openai},
\textsc{Claude\,Opus\,4.8}~\cite{anthropic2025claude},
\textsc{Nemotron-550B}~\cite{blakeman2025nvidia},
\textsc{GLM-5.2}~\cite{zeng2026glm}---and medical-specialized ones:
\textsc{MedGemma-4B}~\cite{sellergren2025medgemma},
\textsc{OpenBioLLM-8B}~\cite{chen2025beyond}, and
\textsc{HuatuoGPT-o1-8B}~\cite{chen2024huatuogpt}. Appendix~\ref{sec:contamination} probes these models for verbatim memorization of
MIMIC-derived text. To control for output-format
priors, a \textbf{few-shot} variant prompts the two strongest general models with
$5$ EarlyDx-style exemplars demonstrating sparse, evidence-supported output.
To ask whether the task requires a generative formulation, a \textbf{supervised}
baseline fine-tunes \textsc{ClinicalBERT}~\cite{huang2019clinicalbert} on the
same split as a multi-label classifier over the  open-vocabulary
titles, emitting its two highest-scoring titles per encounter. We further
evaluate our post-trained model on inputs stripped of all evidence except
demographics and the chief complaint, separating evidence use from label-prior
fitting.

Two \textbf{non-learning baselines} bound what is attainable without diagnostic
inference: a \textbf{frequency prior} predicting the three most common training
diagnoses for every encounter, and a \textbf{retrieval} baseline that embeds each
test encounter with a biomedical retriever~\cite{jin2023medcpt} and copies its
nearest training encounter's diagnoses verbatim---performing no reasoning over the
evidence, it upper-bounds what fitting the site's coding distribution alone can
achieve. As \textbf{post-trained} models we fine-tune \textsc{Qwen3.5-4B} on the EarlyDx
training set (Figure~\ref{fig:finetuning}), supervised with the gold-conditioned
rationale layer of Appendix~\ref{sec:rationale}; hyperparameters are given in
Appendix~\ref{app:config} and all four pipeline prompts in Appendix~\ref{app:prompts}. As a \textbf{human reference} a clinician
independently diagnoses $1{,}000$ random encounters from the same admission-time inputs.

\section{Results}

\begin{table*}[ht]\centering
\caption{Diagnosis performance on EarlyDx by label evidence category. All metrics
are micro-averaged over encounters.  The \textbf{Supported} block is the primary
track; Partial and Secondary are reported separately, and unsupported labels are
excluded from all tracks. \textbf{Bold}: best; \underline{underline}: second best
(per column), among zero-shot, few-shot, and post-trained systems. All pairwise
gains of post-trained over zero-shot models are statistically significant
($p<0.001$, patient-level paired bootstrap). Rows marked $\dagger$ are evaluated on
a subsample rather than the full test split and are not directly comparable.
demo$+$CC: input restricted to demographics and chief complaint.}
\label{tab:full}
\resizebox{0.95\textwidth}{!}{%
\begin{tabular*}{\textwidth}{@{\extracolsep{\fill}} l ccc ccc ccc @{}}
\toprule
& \multicolumn{3}{c}{\textbf{Supported}} & \multicolumn{3}{c}{\textbf{Partial}} & \multicolumn{3}{c}{\textbf{Secondary (S+P)}}\\
\cmidrule(lr){2-4}\cmidrule(lr){5-7}\cmidrule(lr){8-10}
Model & Precision & Recall & F1 & Precision & Recall & F1 & Precision & Recall & F1\\
\midrule
\multicolumn{10}{l}{\emph{Non-learning baselines}}\\
Frequency prior (top-3) & 0.03 & 0.08 & 0.05 & 0.10 & 0.21 & 0.14 & 0.07 & 0.15 & 0.10\\
kNN retrieval           & 0.12 & 0.15 & 0.13 & 0.16 & 0.16 & 0.16 & 0.17 & 0.16 & 0.16\\
\midrule
\multicolumn{10}{l}{\emph{Supervised classifier}}\\
ClinicalBERT  & 0.22 & 0.25 & 0.23 & 0.20 & 0.21 & 0.21 & 0.25 & 0.23 & 0.24\\
\midrule
\multicolumn{10}{l}{\emph{Zero-shot, general}}\\
GPT-5.5            & 0.27 & \underline{0.77} & \underline{0.40} & 0.16 & 0.32 & \underline{0.22} & \underline{0.27} & \underline{0.55} & \underline{0.36}\\
Claude\,Opus\,4.8  & 0.17 & \textbf{0.83} & 0.28 & \underline{0.17} & \textbf{0.57} & \textbf{0.27} & 0.20 & \textbf{0.70} & 0.31\\
GLM-5.2            & 0.26 & 0.69 & 0.37 & 0.11 & 0.24 & 0.15 & 0.23 & 0.47 & 0.31\\
Nemotron-550B      & 0.16 & 0.54 & 0.25 & 0.09 & 0.22 & 0.12 & 0.15 & 0.37 & 0.21\\
\midrule
\multicolumn{10}{l}{\emph{Zero-shot, medical}}\\
MedGemma-4B     & 0.26 & 0.40 & 0.32 & 0.07 & 0.09 & 0.08 & 0.21 & 0.25 & 0.23\\
OpenBioLLM-8B   & 0.10 & 0.53 & 0.17 & 0.08 & \underline{0.34} & 0.13 & 0.11 & 0.43 & 0.17\\
HuatuoGPT-o1-8B & 0.19 & 0.53 & 0.28 & 0.09 & 0.21 & 0.12 & 0.17 & 0.36 & 0.23\\
\midrule
\multicolumn{10}{l}{\emph{Few-shot, general}}\\
GPT-5.5            & \underline{0.30} & 0.59 & \underline{0.40} & 0.08 & 0.09 & 0.08 & 0.24 & 0.32 & 0.27\\
Claude\,Opus\,4.8  & 0.28 & 0.68 & \underline{0.40} & 0.15 & 0.23 & 0.18 & 0.26 & 0.43 & 0.32\\
\midrule
\multicolumn{10}{l}{\emph{Post-trained (ours)}}\\
Qwen3.5-4B (CoT) & \textbf{0.48} & 0.54 & \textbf{0.51} & \textbf{0.27} & 0.26 & \textbf{0.27} & \textbf{0.49} & 0.39 & \textbf{0.43}\\
Qwen3.5-4B (demo$+$CC only) & 0.35 & 0.33 & 0.34 & 0.19 & 0.14 & 0.16 & 0.32 & 0.24 & 0.27\\
\midrule
\multicolumn{10}{l}{\emph{Human reference}}\\
Clinician$^{\dagger}$ & 0.20 & 0.68 & 0.31 & 0.07 & 0.27 & 0.11 & 0.27 & 0.49 & 0.35\\
\bottomrule
\end{tabular*}}
\end{table*}

\subsection{Main Comparison}
\begin{figure}[ht]
  \centering
  \includegraphics[width=0.85\columnwidth]{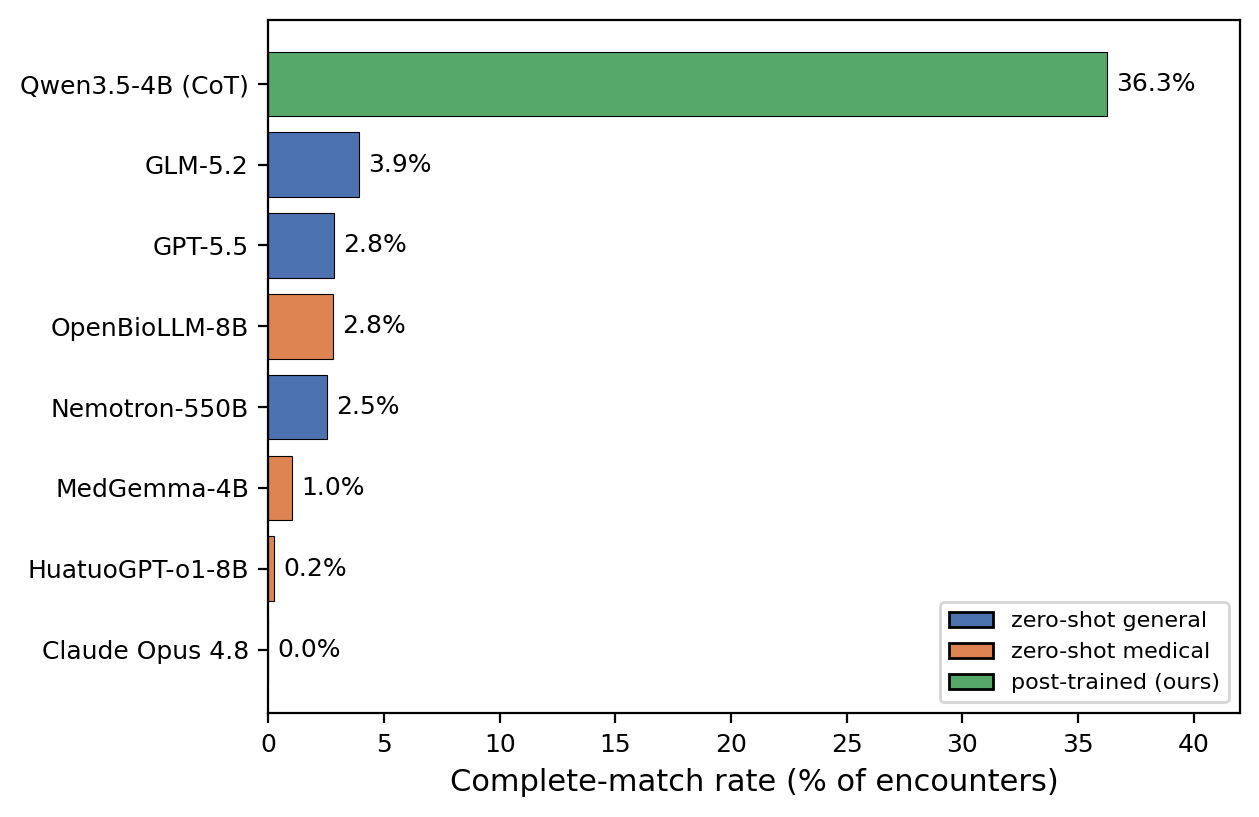}
  \caption{Complete-match rate---the fraction of encounters where the predicted diagnosis set 
  exactly matches the reference. Post-trained models achieve 32--36\%, an order of magnitude 
  higher than any zero-shot model ($<$4\%). The
\emph{Clinician} bar is a human reference on 1000 cases.}
  \label{fig:complete}
\end{figure}

Table~\ref{tab:full} reports diagnosis performance by label evidence category.
Most zero-shot models---general and medical alike---fall below $0.40$ F1,
underscoring the difficulty of early diagnosis. Post-training yields substantial
gains: our fine-tuned model reaches a \emph{supported} (primary-track) F1 of
$0.51$, exceeding the strongest zero-shot baseline by $+0.11$. The bottleneck is
thus not medical knowledge but the capacity to translate it into a calibrated
diagnosis under uncertainty. An input ablation is consistent with this reflecting
evidence use rather than dataset fitting: withholding all admission-time
evidence at inference and retaining only demographics and the chief complaint
lowers supported F1 to $0.34$---the level a frontier model attains on the
complete record. Precision and recall decline in step, so the model is not
merely becoming more conservative, and what remains is what the chief complaint
alone supports; as an inference-time ablation, this figure lower-bounds what
retraining on the reduced input would achieve. The gap widens under exact set
match, where post-trained models recover the full reference set in 32--36\% of
encounters against under 4\% for any zero-shot system
(Figure~\ref{fig:complete}). Finally, because micro-F1 penalises every
prediction beyond the reference set, we also report recall at a matched
prediction budget (R@$k$): frontier models lead at generous budgets while the
post-trained model is competitive at $k{=}1$, the two metrics rewarding coverage
and restraint respectively (Appendix~\ref{sec:recallk}).

\begin{figure}[t]\centering
  \includegraphics[width=\columnwidth]{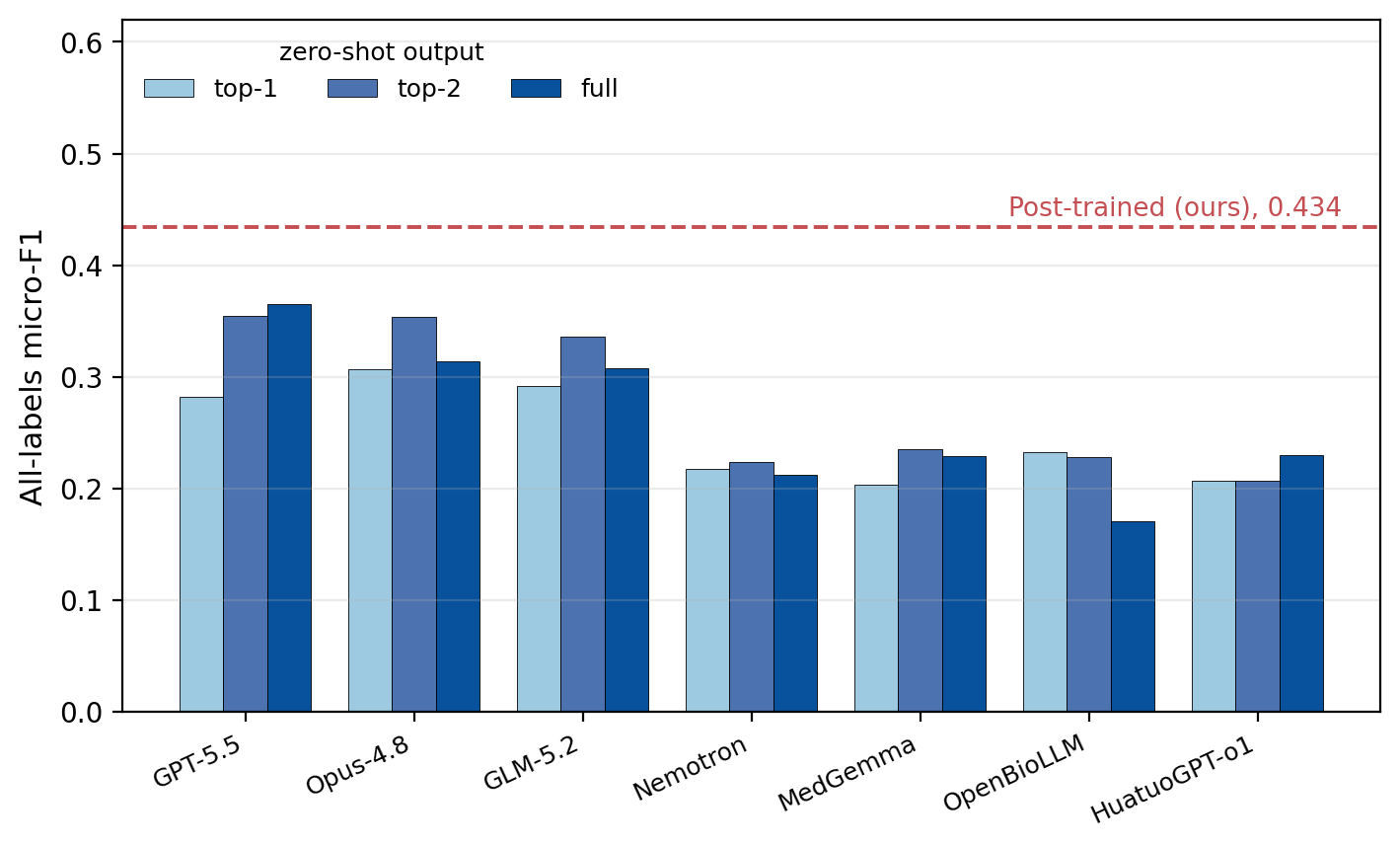}
  \caption{Cardinality-controlled zero-shot baselines (all-labels micro-F1). Each model's
  output is truncated to top-1 / top-2 / full. Even at their best truncation, all general and
  medical zero-shot models stay far below the post-trained models (dotted lines),
  confirming the advantage is not merely from predicting fewer diagnoses.}
  \label{fig:card}
\vspace{-5pt}
\end{figure}

Three controls rule out simpler explanations. A \emph{frequency prior} that always
predicts the three most common training diagnoses attains $0.10$ F1 overall and
$0.05$ on supported labels, so the target is patient-specific rather than
recoverable from the marginal label distribution.  A \emph{retrieval} baseline that
copies the nearest training encounter's diagnoses reaches $0.13$ on supported
labels: fitting the institution's coding distribution recovers barely a quarter of
the post-trained model's performance. A \emph{few-shot} control sharply curbs
over-prediction---predictions per encounter fall from $3.43$ to $2.21$ for GPT-5.5
and $5.78$ to $2.76$ for Claude---yet all-labels F1 stays at $0.27$--$0.32$, so
the sparse output \emph{format} is not what post-training supplies either. Truncating zero-shot output to its top-1 or top-2 diagnoses does not close the gap
either (Figure~\ref{fig:card}), so the advantage is not merely one of
predicting fewer diagnoses.

A discriminative baseline, given a fair configuration, proves competitive with
mid-tier zero-shot LLMs without closing the gap. 
ClinicalBERT attains a
supported F1 of $0.23$: substantially above the frequency prior ($0.05$) and kNN
retrieval ($0.13$), comparable to Nemotron-550B ($0.25$), and closest to the
zero-shot models on the \emph{partial} track ($0.21$ against $0.27$ for the best
of them), yet well short of the post-trained generative model ($0.51$).
Finally, the human clinician attains the highest \emph{supported} recall ($0.68$)
but an all-labels F1 comparable to GPT-5.5: an experienced clinician also
enumerates a broad differential that the sparse reference penalizes, so
over-prediction is not an LLM-specific artifact. That a human expert working from
serialized results alone---rather than a live encounter with history-taking and
examination---scores in this range further suggests our text-rendered inputs bound
the achievable ceiling.


\subsection{Extraction versus Inference}
\label{sec:extraction}
Any benchmark built over clinical notes faces a construct-validity threat: if
reference diagnoses are frequently named outright in the input, high scores may
reflect text extraction rather than diagnostic inference. We test this by
partitioning gold labels on recoverability by string matching alone. A supported
diagnosis is \emph{explicit} if its full title, or every content word, occurs
verbatim in the input, and \emph{implicit} otherwise. Under this
operationalization $43\%$ of supported labels are explicit; the remaining $57\%$
are never stated and must be inferred.

\begin{figure}[t]
  \centering
  \includegraphics[width=0.9\columnwidth]{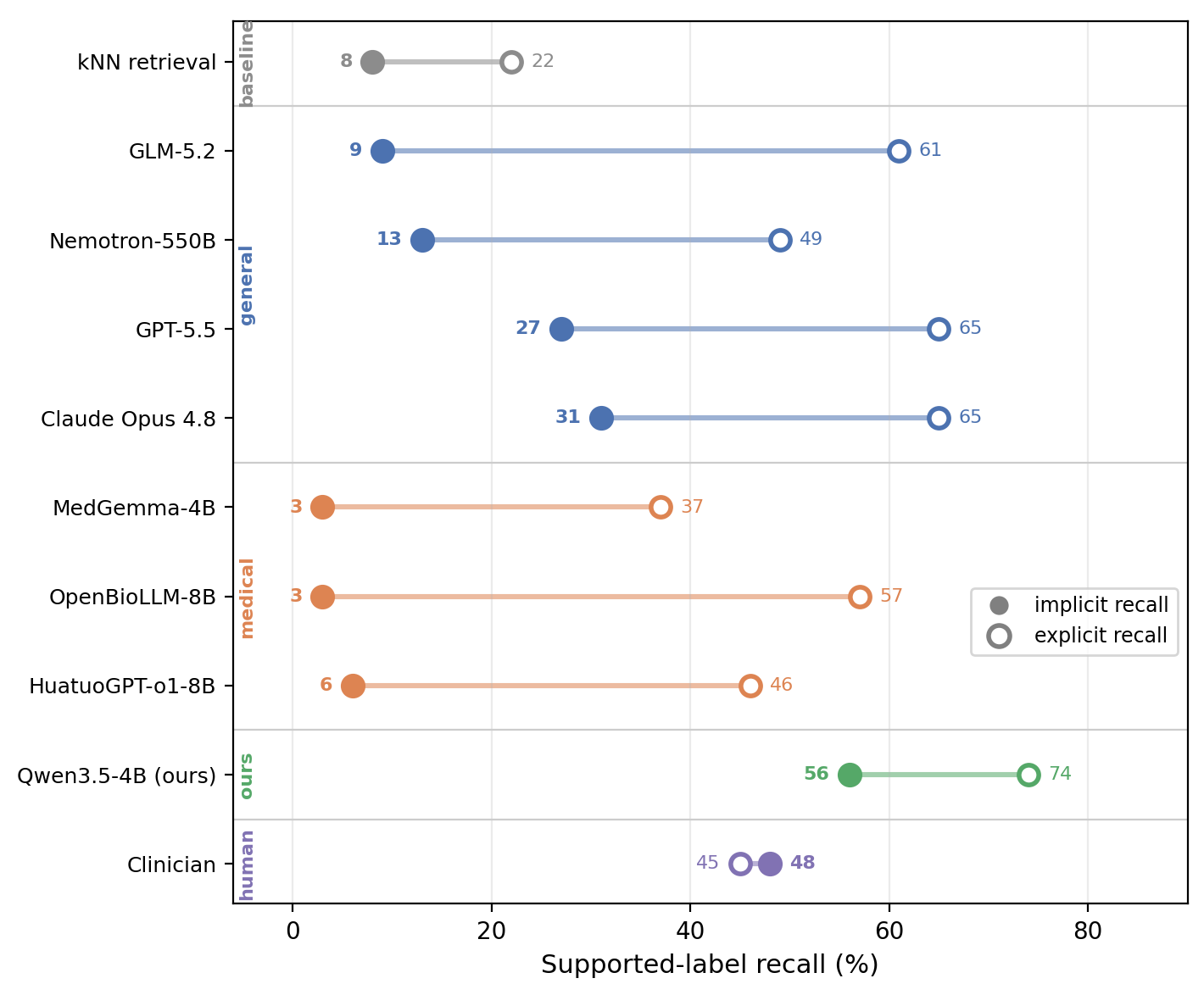}
  \caption{Recall on \emph{explicit} ($\circ$; the diagnosis appears verbatim in
  the input) versus \emph{implicit} ($\bullet$; it must be inferred) supported
  labels at $W{=}0$; only $43\%$ are explicit. Zero-shot models---medical ones
  most severely---collapse on the implicit majority, whereas post-training
  recovers inference at comparable explicit recall.}
\vspace{-5pt}
  \label{fig:extract}
\end{figure}

Figure~\ref{fig:extract} reports recall on each subset.\footnote{Recall is
reported because the partition is defined over gold labels, whereas precision is
defined over predictions and admits no corresponding split. Explicitness is
detected lexically, so paraphrased mentions count as implicit and $43\%$ is a
lower bound on the extractable fraction.} Every system performs worse on implicit
labels, confirming the partition isolates genuinely harder instances. The
informative result is a dissociation between families: zero-shot models are
competitive on explicit labels ($37$--$65\%$) yet collapse on implicit ones,
recovering $9$--$31\%$ (general) and $3$--$6\%$ (medical), so their apparent
competence is largely attributable to extraction. Post-training reverses this
profile---Qwen3.5-4B attains comparable explicit recall ($74\%$) while recovering
$56\%$ of implicit labels.

One alternative explanation requires exclusion. Implicit labels are unstated but
need not require reasoning: they may be predictable from coding regularities, as
when a comorbidity is habitually coded alongside a presenting condition. The
retrieval baseline isolates this, encoding nothing but the dataset prior, and
recovers $22\%$ of explicit but only $8\%$ of implicit labels---below every
zero-shot system on the implicit subset. Dataset priors are thus least
informative exactly where inference is required.

The human reference breaks the pattern entirely: the clinician exhibits no
explicit--implicit gap ($45\%$ vs.\ $48\%$). Whether a diagnosis is spelled out is
a property of the \emph{documentation}, not of the diagnostic problem, and an
expert reasoning from the underlying findings is indifferent to it. The gap exhibited by every
automated system measures residual reliance on surface cues.

Together these results attribute the effect of post-training to inference over
evidence the record leaves unstated, rather than to more effective copying or
distributional fitting---and domain-specific pre-training alone does not confer
it. They also bound the extraction confound for the benchmark: because implicit
labels form the majority and determine rankings, EarlyDx measures synthesis under
uncertainty rather than retrieval of stated text.

\subsection{Direct Evidence Yields More Accurate Diagnosis}
\label{sec:direct}
Performance separates sharply across the evidence boundary
(Table~\ref{tab:full}): for our best model, F1 falls from $0.51$ on supported to
$0.27$ on partially supported labels, and the pattern holds for every baseline.
Supported diagnoses can be traced to a specific early finding, whereas partially
supported ones rest on indirect cues with no confirmatory evidence at admission;
the latter therefore govern the predictability ceiling of early diagnosis.
Appendix~\ref{app:cot-examples} gives representative chain-of-thought traces for
one case of each type. Appendices~\ref{app:modality} and~\ref{app:ablation} localize which modality each
diagnosis depends on, by rationale attribution and by input ablation respectively.

\subsection{Output Cardinality and the Recorded Reference}
\begin{figure}[t]
  \centering
  \includegraphics[width=0.85\columnwidth]{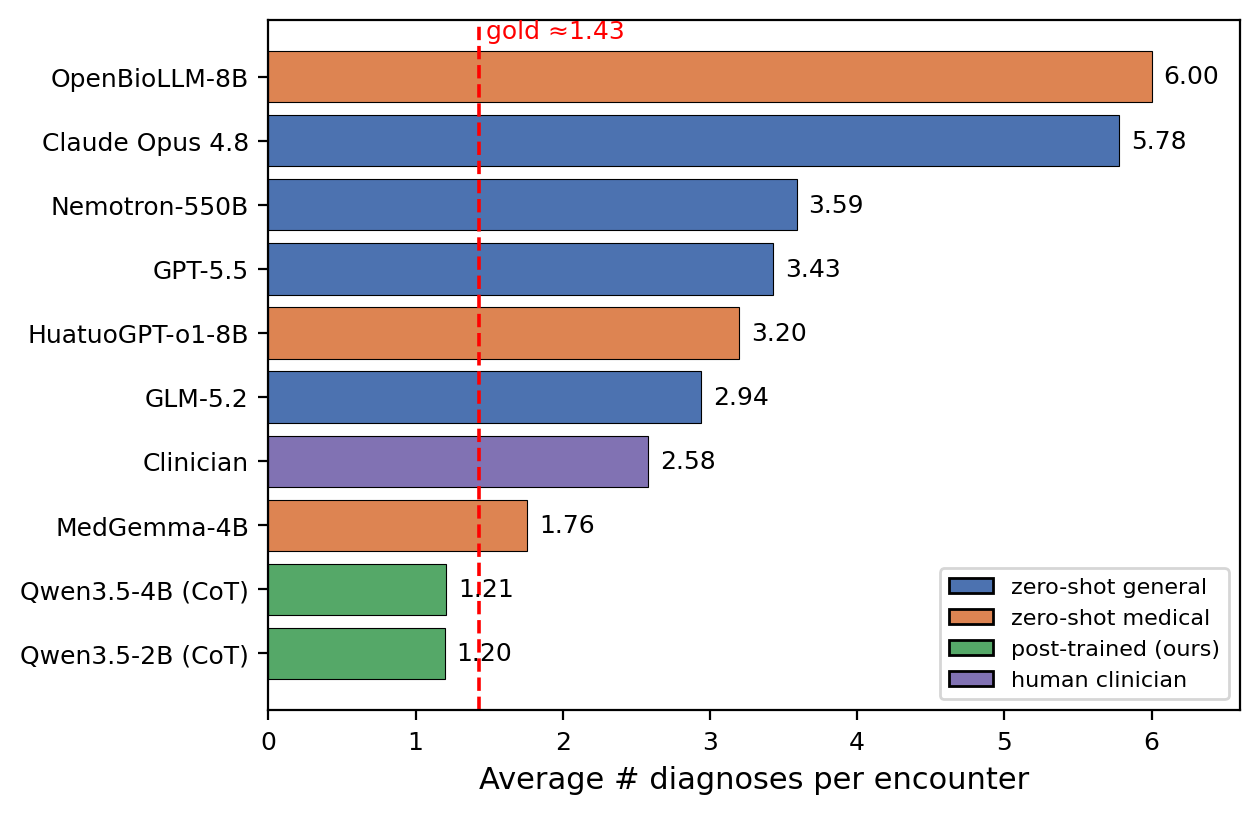}
  \caption{Average number of predicted diagnoses per encounter. Zero-shot models
over-predict relative to the reference ($\approx$1.4, dashed line); post-trained
models match it; a human clinician ($2.6$) lies in between, listing a broader
differential.}
  \label{fig:avgdx}
\end{figure}

Across systems the dominant discrepancy against the reference is one of
cardinality: zero-shot models emit $1.76$--$6.0$ diagnoses per encounter against a
reference of roughly $1.4$ (Fig.~\ref{fig:avgdx}), yielding high recall at low
precision. Claude Opus 4.8 exemplifies the pattern, attaining the best recall in
every scope ($0.83$ supported) at among the lowest precision ($0.17$). Our
post-trained model predicts ${\sim}1.2$ diagnoses, matching the reference
distribution, and leads on precision and F1 throughout.

The human reference indicates this is not a model deficit. Working from the same
record, an experienced clinician also over-lists relative to the reference
($2.58$ per encounter) and shows the zero-shot profile---high supported recall
($0.68$) at $0.20$ precision. Breadth under incomplete information is thus a
property of diagnostic reasoning where the evidence does not yet determine a
single answer and expert practice is to enumerate what must be excluded.

Two limitations of the evaluation produce this mismatch. The reference records the
diagnoses ultimately coded for billing, so a differential entertained at admission
and subsequently excluded leaves no trace. And the evidence captures what was
written rather than what was observed---the patient's appearance, respiratory
effort, physical examination, and the course of the encounter are largely absent,
yet these are what narrow a differential at the bedside. A clinician reasoning
from the record alone is deprived of them equally, attaining the highest supported
recall at only $0.31$ F1: the evidence suffices to identify the condition but not
to exclude the alternatives. We therefore report the cardinality gap as a property
of the evaluation rather than as evidence that broader outputs are clinically
unwarranted.

\subsection{Risk-Weighted Evaluation: Time-Critical Diagnoses}
\label{sec:critical}
\begin{figure}[t]
  \centering
  \includegraphics[width=0.8\columnwidth]{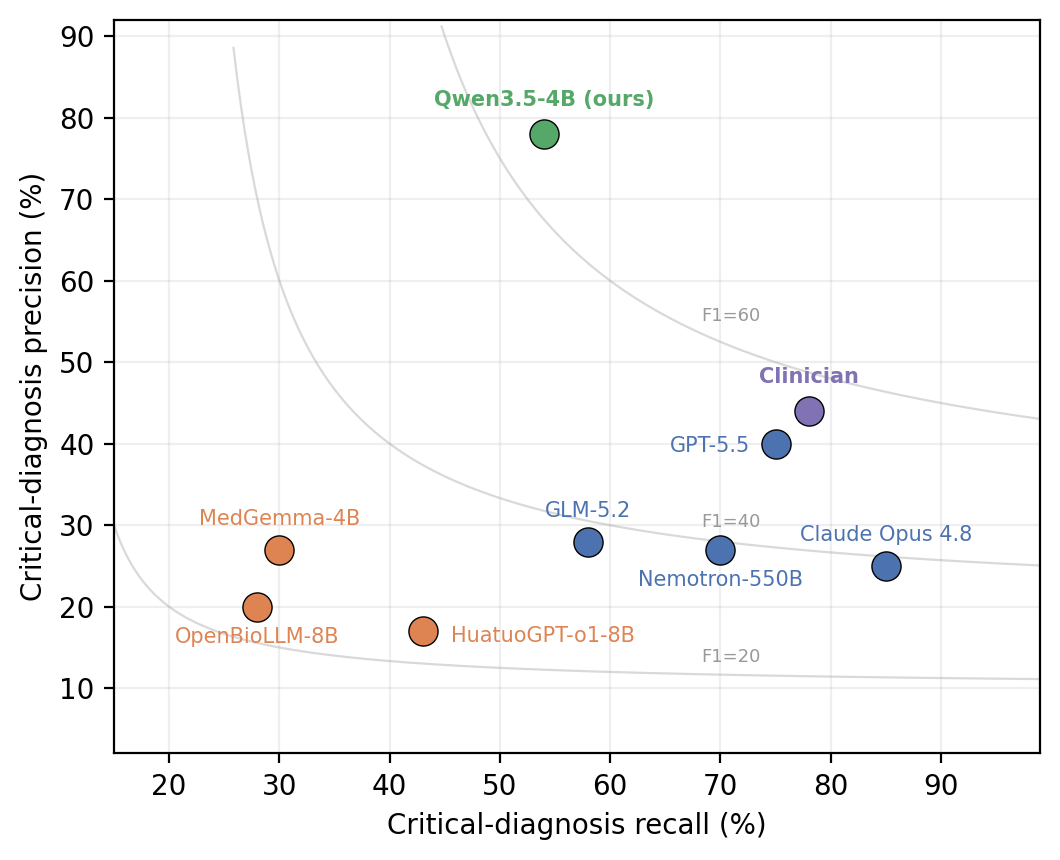}
  \caption{Operating points on six time-critical diagnoses at $W{=}0$; grey curves
  are F1 isolines. The clinician ($78\%$ recall, $44\%$ precision) marks the
  conventional clinical trade-off. Zero-shot models match or exceed that
  sensitivity only at $25$--$28\%$ precision, medical-specialized models are
  dominated on both axes, and post-training attains the highest precision at
  reduced sensitivity---no system falls near the clinician's operating region.
  }
  \label{fig:critical}
\end{figure}

The metrics so far treat all diagnoses as exchangeable, whereas the loss function
of emergency care is asymmetric: omitting a myocardial infarction or sepsis at
admission carries a cost that omitting a stable comorbidity does not. We therefore
re-evaluate the same predictions over six time-critical conditions---myocardial
infarction, sepsis, intracranial hemorrhage, pulmonary embolism, stroke, and
gastrointestinal bleeding---computing encounter-level recall and precision for
every system, the clinician included
(Figure~\ref{fig:critical}).

The clinician provides a reference operating point---$78\%$ recall at $44\%$
precision, i.e.\ $1.8\times$ over-flagging---reflecting an expert who widens the
threshold for dangerous conditions, but selectively. Every automated system
departs from it along one of two axes. Zero-shot models match or exceed the
clinician's sensitivity (Claude $85\%$, GPT-5.5 $75\%$, Nemotron-550B $70\%$) at
$25$--$28\%$ precision; Claude raises a critical-diagnosis alert in $46$ of $200$
random encounters in which $14$ are recorded. Medical-specialized models are
dominated on both axes. Our post-trained model shows the opposite profile:
precision is highest by a wide margin ($78\%$), while $54\%$ recall appears to
fall short of the clinician on the axis along which errors are least recoverable.

These operating points suggest a frontier that no evaluated system appears to
reach: within the precision range we observe, sensitivity adequate for
time-critical conditions comes at precision well below clinical tolerance, and
precision within that tolerance at the cost of sensitivity. Since the clinician
and two zero-shot systems are scored on subsamples rather than the full test split, these comparisons are indicative rather than
definitive. The pattern is nonetheless invisible to aggregate F1, where our
model ranks 1st, indicating that a risk-weighted view carries information the
headline metric does not. It also delimits present formulation: restricted to
critical labels the verifier classes as \emph{supported}, our model leads all
systems ($85\%$ recall against $78\%$ for Nemotron-550B), which locates much of
its aggregate shortfall in references whose confirmation is still pending at
$t_0$. Supervision by \emph{recorded} diagnoses rewards silence wherever
confirmation is absent, while admission-time practice requires that unconfirmed
dangers be raised. Extending EarlyDx to graded predictions---diagnoses marked as
suspected pending confirmation, and scored under a loss that credits them---would
render this capability directly optimizable rather than merely observable.

\section{Discussion and Conclusion}
We introduced EarlyDx, a large-scale benchmark for open-ended early diagnosis
built from MIMIC-IV admissions, in which every label is verified against the
evidence actually available at admission and scoring uses an open-weight,
version-pinned semantic judge. The central finding is negative but useful:
zero-shot models mostly extract diagnoses the record already names, and collapse
when one has to be inferred. Post-training narrows that gap without closing it,
and on time-critical conditions no system reaches a clinician's balance of
sensitivity and precision.
Two features of the evaluation limit what these results can show. The reference
lists only the diagnoses finally coded, so a differential a clinician rightly
considered and then ruled out counts as an error. And the record captures what was
written, not what was seen---the patient's appearance, the examination, how the
encounter unfolded. Human and model alike work from less than the bedside offers,
so reasoning faithfully under uncertainty is penalized, and agreement with this
reference should not be read as clinical adequacy. Letting systems mark diagnoses
as suspected pending confirmation would remove that confound, and we see it as the
most valuable next step. Other limitations remain: single-center data,
text-rendered signals, and a rationale layer we release only as an exploratory
resource (Appendix~\ref{app:cot_review}). Code, data access terms, ethics, and a
datasheet are in
Appendices~\ref{app:availability}--\ref{app:datasheet}.

\clearpage

\bibliographystyle{ACM-Reference-Format}
\bibliography{sample-base}

\appendix
\section{A cardinality-tolerant view: recall at a matched budget}
\label{sec:recallk}

Micro-F1 penalises every prediction beyond the reference set. With $1.43$
reference diagnoses per encounter, a system emitting $5.78$ of them---as Claude
Opus 4.8 does---faces an arithmetic precision ceiling of $1.43/5.78 \approx
0.25$, so its F1 is governed by output cardinality rather than diagnostic
quality. We therefore report a second primary metric that is insensitive to how
much a system chooses to say.

\paragraph{Metric.}
Following the top-$k$ differential-diagnosis accuracy of DDxPlus~\cite{fansi2022ddxplus} and the AMIE~\cite{tu2024towards}
line of work, we truncate each system to its first $k$ predictions and measure
supported-label recall, R@$k$. The budget is identical across systems, so
listing more confers no advantage. Since no system emits confidence scores,
truncation follows the model's own output order.

\begin{figure}[t]
\centering
\includegraphics[width=\columnwidth]{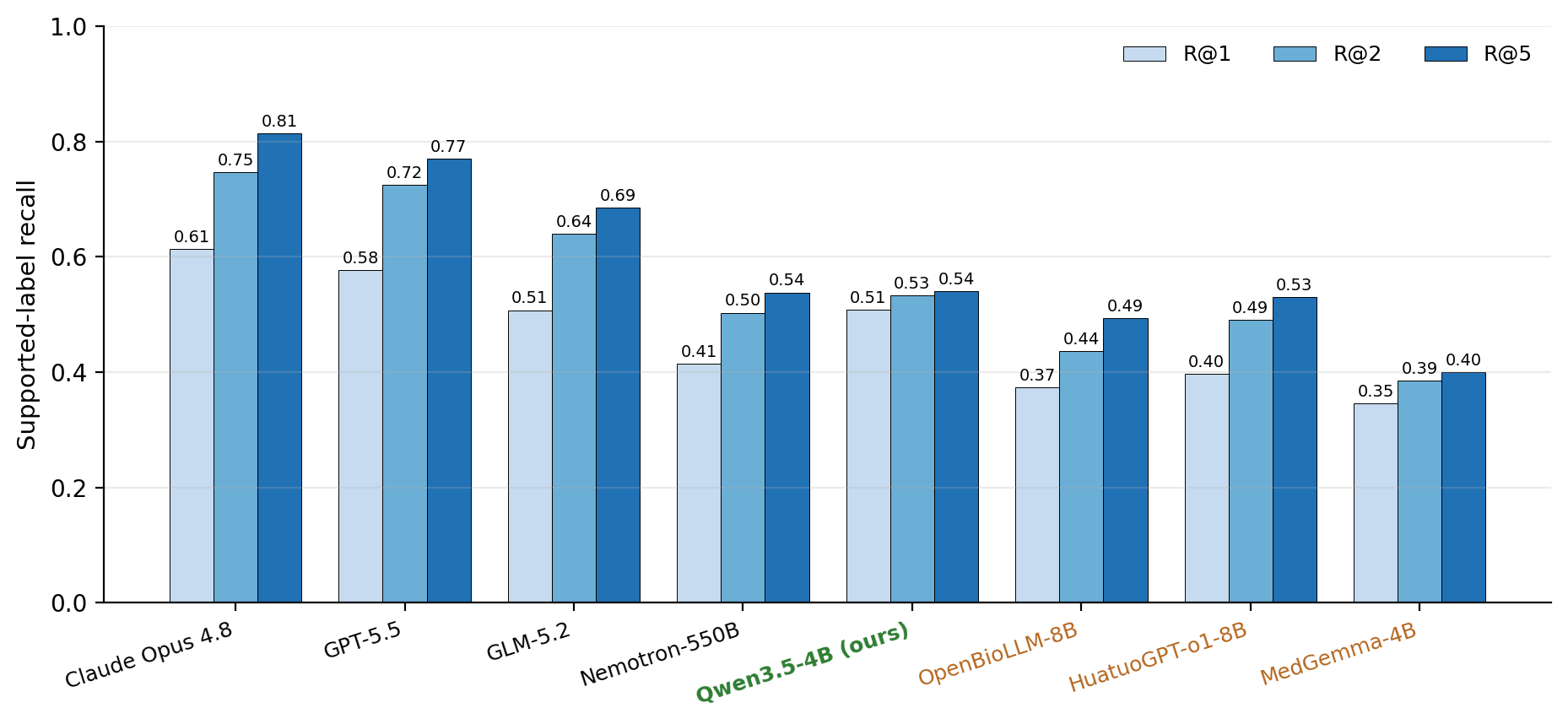}
\caption{Supported-label recall at a matched prediction budget. Every system is
truncated to its first $k$ predictions, so none benefits from listing more.}
\label{fig:recallk}
\end{figure}

\paragraph{Result.}
\emph{The post-trained model does not lead under this metric}
(Figure~\ref{fig:recallk}). At $k{=}5$ the ordering tracks output cardinality:
Claude Opus 4.8 attains $0.81$ and GPT-5.5 $0.77$, against $0.54$ for our 4B
model. The comparison inverts at $k{=}1$, where each system may name a single
diagnosis: our model reaches $0.51$, matching GLM-5.2 and exceeding
Nemotron-550B by $0.10$ even though both converge to $0.54$ at $k{=}5$. The
post-trained curve is nearly flat ($0.51 \rightarrow 0.54$)---the model emits
$1.2$ diagnoses on average and cannot exploit a larger budget---whereas the
frontier models gain $0.16$--$0.20$ from $k{=}1$ to $k{=}5$, indicating that
their correct answers lie predominantly in later positions.

\paragraph{Interpretation.}
The two metrics disagree because they reward different behaviours: recall at a
generous budget rewards enumerating a broad differential, while micro-F1 and
complete-match reward committing to a small, defensible set. Post-training does
not expand what a model can eventually retrieve---its R@$\infty$ equals
Nemotron-550B's---but relocates the correct diagnosis to the first position and
suppresses the remainder. Neither behaviour dominates: a triage display
surfacing one working diagnosis favours the former, a differential worksheet the
latter. We therefore report both metrics rather than designating a single
ranking, and treat their divergence as a substantive property of admission-time
diagnosis rather than an evaluation artefact.

\section{Robustness to the Evidence-Class Boundary}
\label{app:boundary}
The primary track rests on a single LLM auditor decision---whether a label's
evidence is \emph{supported} or merely \emph{partial}---so we ask whether that
decision, rather than model behaviour, determines the reported ranking.

\noindent\textbf{Independent re-audit.} Four experts independently re-rated a random sample of $1{,}000$ test
encounters. At least three of the four confirm the auditor's assignment in $76\%$
of cases (Table~\ref{tab:reaudit}).

\begin{table}[h]
\centering\small
\caption{Re-audit of a random sample of $1{,}000$ test encounters. Agreement is
computed against the LLM auditor; the last column averages each expert's $\kappa$
with the other three.}
\label{tab:reaudit}
\begin{tabular}{lcc}
\toprule
Auditor & agreement & mean $\kappa$ vs.\ peers \\
\midrule
Expert 1 & 75\% & 0.63 \\
Expert 2 & 67\% & 0.51 \\
Expert 3 & 76\% & 0.64 \\
Expert 4 & 71\% & 0.57 \\
\bottomrule
\end{tabular}
\end{table}

\noindent\textbf{Re-scoring under a stricter gold standard.}
Table~\ref{tab:boundary} scores every system on these encounters twice: against
the LLM auditor gold, and against gold restricted to supported labels confirmed by
${\geq}3$ of $4$ human auditors.

\begin{table}[h]
\centering\small
\caption{Micro-F1 on the $1{,}000$ re-audited test encounters under the original
and the auditor-confirmed gold standard.}
\label{tab:boundary}
\begin{tabular}{lcc}
\toprule
Model & verifier gold & ${\geq}3/4$ confirmed \\
\midrule
Qwen3.5-4B (ours) & \textbf{0.563} & \textbf{0.697} \\
GPT-5.5           & 0.375 & 0.430 \\
Nemotron-550B     & 0.334 & 0.342 \\
GLM-5.2           & 0.253 & 0.297 \\
Claude Opus 4.8   & 0.216 & 0.253 \\
\bottomrule
\end{tabular}
\end{table}

The ranking is invariant to the choice of gold standard, so it is not an artifact
of where the verifier draws the supported/partial line; the margin of the
post-trained model over the strongest zero-shot system in fact widens from
$+0.188$ to $+0.267$. The differential effect of filtering is itself informative:
largest for the post-trained model ($+0.134$) and negligible for Nemotron-550B
($+0.008$), it indicates that post-training concentrates predictions on labels
whose evidence multiple independent auditors endorse, whereas a larger share of
the zero-shot models' matches falls on disputed labels. Inter-expert agreement is
moderate ($\kappa{=}0.51$--$0.64$), reflecting genuine clinical ambiguity at the
supported/partial boundary; the invariance of the ranking under both gold
standards is what makes the primary track robust to it.

\section{Robustness to the Treatment of Partially Supported Labels}
\label{app:ignore}
The primary track scores against supported labels alone, so a prediction matching
a \emph{partially supported} label is counted as a false positive despite
appearing in the record. Because the evidence class is not observable to the
model, this convention could disadvantage systems that emit broader
differentials---precisely the dimension along which the compared systems differ
most.

We therefore recompute every system under an \emph{ignore} variant, in which
predictions matching partially supported labels are withheld from the precision
denominator rather than penalized, following the treatment of unjudged documents
in retrieval evaluation. Recall is invariant by construction, since neither the
matched count nor the reference set is altered.

Rankings are identical under both treatments and the shifts are uniformly small
(Table~\ref{tab:ignore}), so the convention does not account for the observed
ordering. Their direction runs contrary to the concern motivating the analysis:
the largest gains accrue to the post-trained models and among the smallest to the
zero-shot system with the highest recall and broadest output. Predictions from
broadly predicting systems are, in the main, absent from the record altogether
rather than classed as partially supported, so excusing partial matches does
little to recover their precision.

\begin{table}[h]\centering\small
\caption{Change in primary-track precision and F1 under the \emph{ignore}
variant, in which predictions matching partially supported labels are withheld
from the precision denominator. Recall is invariant by construction and is given
for reference.}
\label{tab:ignore}
\begin{tabular}{lccc}
\toprule
Model & R & $\Delta$P & $\Delta$F1 \\
\midrule
Qwen3.5-4B (CoT) & $0.575$ & $+0.025$ & $+0.013$ \\
GPT-5.5          & $0.761$ & $+0.006$ & $+0.007$ \\
Nemotron-550B    & $0.654$ & $+0.004$ & $+0.005$ \\
GLM-5.2          & $0.699$ & $+0.003$ & $+0.004$ \\
Claude Opus 4.8  & $0.832$ & $+0.004$ & $+0.006$ \\
\bottomrule
\end{tabular}
\end{table}

\section{Sensitivity to Timestamp Semantics}
\label{app:store}
Most MIMIC-IV modalities carry two timestamps: \texttt{charttime}, the time the
event occurred or the specimen was drawn, and \texttt{storetime}, the time the
result became readable in the record. The main benchmark filters on
\texttt{charttime}, the more inclusive convention: a CT performed before $t_0$ is
admitted even if its report was finalized afterwards. Because this could admit
information not yet clinically available at the admission decision, we repeat the
$W{=}0$ evaluation under \texttt{storetime} filtering for laboratory results and
radiology reports.

The stricter convention removes $19.3\%$ of radiology reports and $2.7\%$ of
laboratory results, altering the input for $25.9\%$ of encounters and the model's
predictions for $11.7\%$. Primary-track F1 falls from $0.546$ to $0.512$
($-0.034$; precision $0.519\!\to\!0.487$, recall $0.575\!\to\!0.538$), leaving the
post-trained model well above the strongest zero-shot baseline ($0.385$). Results
filtered on event time should therefore be read as a modest upper bound on what
was strictly readable at $t_0$---the convention bounds the reported figures
without affecting the comparison.

\begin{table}[h]\centering\small
\caption{Sensitivity to timestamp semantics at $W{=}0$ (primary track}
\label{tab:store}
\begin{tabular}{lccc}
\toprule
Filter & Precision & Recall & F1 \\
\midrule
\texttt{charttime} (main) & $0.519$ & $0.575$ & $0.546$ \\
\texttt{storetime}        & $0.487$ & $0.538$ & $0.512$ \\
\midrule
$\Delta$                  & $-0.032$ & $-0.037$ & $-0.034$ \\
\bottomrule
\end{tabular}
\end{table}

\section{Contamination}
\label{sec:contamination}
MIMIC-derived text circulates widely, so any evaluation of frontier LLMs on this
corpus must establish whether the models have prior exposure to it. We probe for
verbatim memorization: held-out radiology reports are truncated at the midpoint of
their findings section and each model is prompted to continue them, with the
continuation scored against both the withheld text and a randomly drawn report
from the same corpus. The control absorbs the templated phrasing characteristic of
radiology reporting, so only agreement in excess of it is evidence of recall.

The open-weight baseline yields no exact $10$-gram match with the withheld
continuation, its longest common run scarcely exceeding control
(Table~\ref{tab:contam}). Both proprietary models show a weak but non-zero signal,
matching a small fraction of $10$-grams exactly where the control matches none,
with the longest single run extending to twelve tokens.

This is insufficient to establish memorization---a twelve-token run remains
compatible with standard reporting templates, and the proprietary samples are
small---but sufficient to establish that prior exposure cannot be \emph{excluded}
for those models; nor can any black-box probe exclude exposure that shaped model
behaviour without inducing verbatim recall. Two observations nonetheless bound the
practical concern. First, the model for which no signal is detected is not the
weakest zero-shot system, so the observed ranking cannot be attributed to
contamination. Second, memorization predicts recovery of diagnoses the record does
not state, whereas zero-shot recall on implicit labels
collapses (Section~\ref{sec:extraction})---precisely the inverse of the predicted
pattern.

\begin{table}[t]\centering\small
\caption{Verbatim-continuation probe on $200$ reports. \emph{True} scores the
continuation against the withheld text, \emph{control} against a random report
from the same corpus; only excess over control is evidence of recall. Longest run
is averaged over reports.}
\label{tab:contam}
\begin{tabular}{lcccc}
\toprule
 & \multicolumn{2}{c}{Longest run (mean)} & \multicolumn{2}{c}{Exact 10-gram} \\
\cmidrule(lr){2-3}\cmidrule(lr){4-5}
Model & True & Ctrl & True & Ctrl \\
\midrule
Nemotron-550B (open) & $2.9$ & $1.6$ & $0.0\%$ & $0.0\%$ \\
GPT-5.5              & $3.2$ & $1.3$ & $6.3\%$ & $0.0\%$ \\
Claude Opus 4.8      & $4.2$ & $1.7$ & $3.2\%$ & $0.0\%$ \\
\bottomrule
\end{tabular}
\end{table}

\section{Subgroup Performance}
\label{app:subgroup}
This appendix reports both the \emph{distribution} of evidence classes across
patient subgroups, referenced from Section~\ref{sec:subgroup}, and the resulting
\emph{performance} by subgroup.

\begin{figure}[t]
  \centering
  \includegraphics[width=\columnwidth]{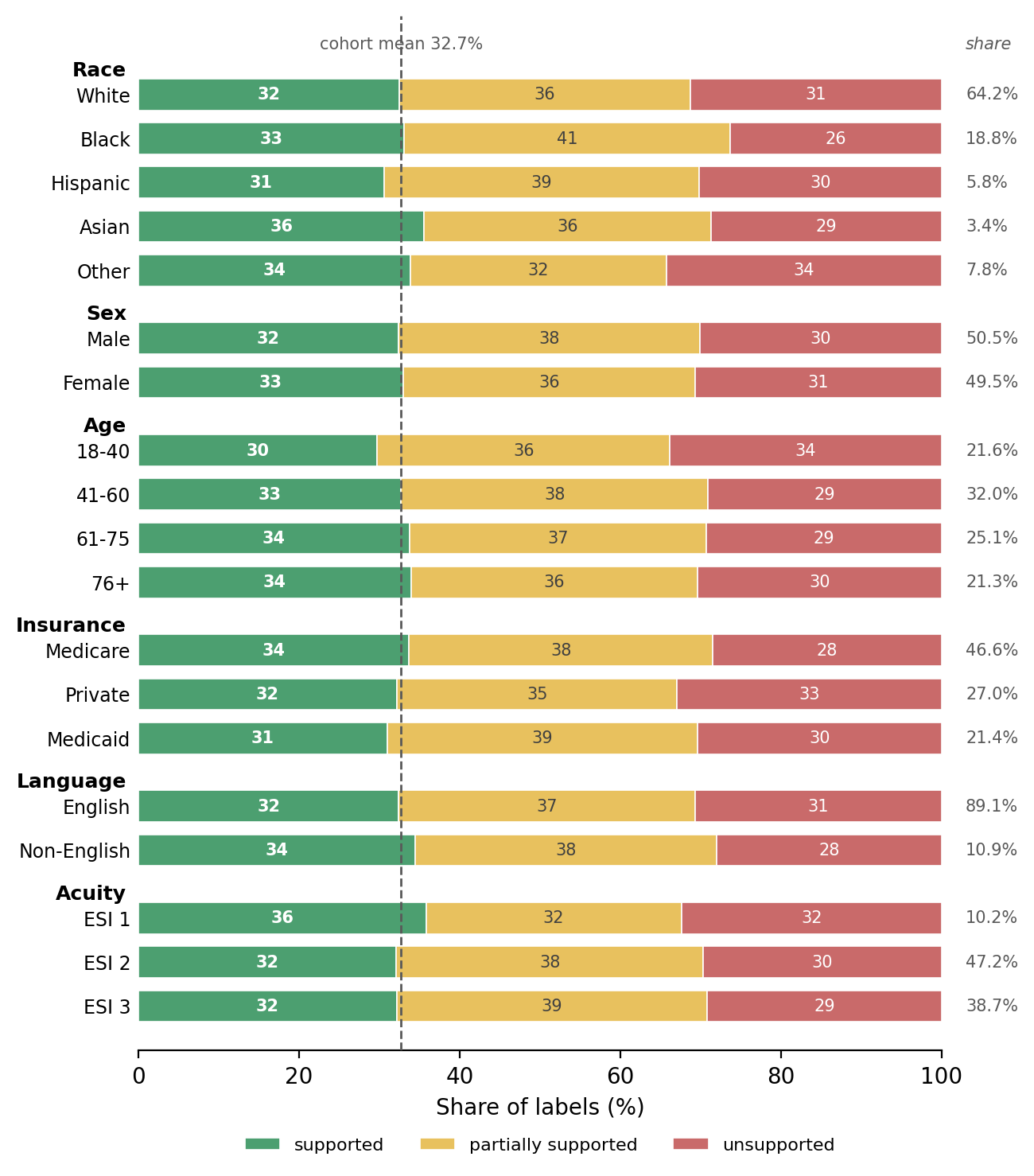}
  \caption{Cohort composition and evidence availability by subgroup at $W{=}0$.
  Bars give each group's label distribution over evidence classes; the right-hand
  column gives its share of encounters. Supported rates cluster tightly around the
  cohort mean (dashed line), spanning at most $5$ points across any grouping and
  none by sex ($p{=}0.32$).}
  \label{fig:cohort}
\end{figure}

\noindent\textbf{Evidence-class distribution.} Figure~\ref{fig:cohort} gives each
group's label distribution over evidence classes alongside its share of
encounters. The cohort is adult and skewed toward higher acuity---$57\%$ of
encounters are triaged ESI~1--2 and $53\%$ arrive by ambulance---as expected of ED
visits ending in admission; race and language distributions reflect the source
institution's catchment and are not nationally representative. Supported rates
cluster tightly around the cohort mean of $32.7\%$, spanning at most $5$
percentage points across any grouping and none by sex ($p{=}0.32$). What residual
variation there is follows clinical rather than demographic axes---ESI~1
presentations have the highest supported rate, consistent with a more extensive
workup---so evidence availability varies more with how long one observes than with
who the patient is.

\noindent\textbf{Performance.} Table~\ref{tab:subgroup_perf} reports primary-track
(supported) F1 by subgroup for four systems spanning three families: our
post-trained model, two frontier zero-shot models, and a medical-specialized
model. Scores are computed on encounters with a cached judge decision
($2{,}247$--$2{,}822$ per system), so estimates for the least represented groups
(Asian, Hispanic, non-English) rest on fewer than $150$ encounters and are
correspondingly noisy.

Flat evidence availability does not imply flat performance. Two patterns hold
across all four systems. Performance is approximately flat across race, sex, and
language: the spread across racial groups is $0.067$ F1 for our model and $0.020$
for OpenBio, of the same order as the sampling noise implied by the smaller
strata. Performance instead declines monotonically with patient age---$0.802$ to
$0.718$ (ours), $0.492$ to $0.346$ (Nemotron), $0.246$ to $0.157$
(OpenBio)---with the Medicare stratum, largely coterminous with the oldest group,
showing the same effect. Since the decline appears even in models never trained on
EarlyDx, it is unlikely to originate in our supervision; older patients carry more
comorbidities and more labels per encounter ($1.72$ versus $1.42$ for the youngest
group), making exact set agreement harder. We report the trend rather than adjust
for it, and recommend that age-stratified results accompany any use of this
benchmark.

\begin{table}[t]\centering
\caption{Primary-track (supported) F1 by subgroup at $W{=}0$. Groups with fewer
than $40$ scored encounters are omitted.}
\label{tab:subgroup_perf}
\resizebox{\columnwidth}{!}{%
\begin{tabular}{llcccc}
\toprule
 & Group & Post-trained & \multicolumn{2}{c}{Zero-shot} & Medical \\
\cmidrule(lr){3-3}\cmidrule(lr){4-5}\cmidrule(lr){6-6}
 & & Qwen3.5-4B & Nemotron & GLM-5.2 & OpenBio \\
\midrule
\multirow{5}{*}{Race} & White & $0.757$ & $0.385$ & $0.347$ & $0.176$\\
 & Black & $0.791$ & $0.374$ & $0.372$ & $0.171$\\
 & Hispanic & $0.724$ & $0.316$ & $0.347$ & $0.180$\\
 & Asian & $0.756$ & $0.372$ & $0.365$ & $0.191$\\
 & Other & $0.753$ & $0.424$ & $0.403$ & $0.186$\\
\midrule
\multirow{2}{*}{Sex} & Male & $0.778$ & $0.389$ & $0.357$ & $0.182$\\
 & Female & $0.746$ & $0.375$ & $0.356$ & $0.169$\\
\midrule
\multirow{4}{*}{Age} & 18--40 & $0.802$ & $0.492$ & $0.476$ & $0.246$\\
 & 41--60 & $0.785$ & $0.375$ & $0.374$ & $0.176$\\
 & 61--75 & $0.741$ & $0.355$ & $0.314$ & $0.154$\\
 & 76+ & $0.718$ & $0.346$ & $0.295$ & $0.157$\\
\midrule
\multirow{3}{*}{Insurance} & Medicare & $0.728$ & $0.335$ & $0.299$ & $0.155$\\
 & Private & $0.762$ & $0.454$ & $0.426$ & $0.202$\\
 & Medicaid & $0.826$ & $0.401$ & $0.393$ & $0.184$\\
\midrule
\multirow{2}{*}{Language} & English & $0.766$ & $0.384$ & $0.360$ & $0.176$\\
 & Non-English & $0.739$ & $0.365$ & $0.319$ & $0.175$\\
\midrule
\multirow{3}{*}{Acuity} & ESI 1 & $0.730$ & $0.351$ & $0.318$ & $0.175$\\
 & ESI 2 & $0.759$ & $0.360$ & $0.338$ & $0.168$\\
 & ESI 3 & $0.777$ & $0.420$ & $0.392$ & $0.182$\\
\bottomrule
\end{tabular}}
\end{table}

\section{Optional Gold-Conditioned Rationale Annotations}
\label{sec:rationale}
Beyond the verified labels, EarlyDx provides an optional rationale layer whose
purpose is \emph{clinical interpretability}: a rationale that grounds each
diagnosis in a specific admission-time finding---a lab value, an imaging finding,
a vital sign---makes a prediction auditable rather than opaque, and is a
prerequisite for clinical trust. Given the admission-time input $x$ and the
retained reference diagnoses $Y$, a teacher LLM
(MiMo-V2.5~\cite{coreteam2026mimov2flashtechnicalreport}) produces a response of
the form
\texttt{<think>}\,\emph{reasoning}\,\texttt{</think>}\,%
\texttt{<answer>}\,\emph{diagnoses}\,\texttt{</answer>}.
Because the reasoning is conditioned on the gold diagnoses, it is not a
ground-truth reasoning trace but a post-hoc explanation of known labels,
following the rationalization
paradigm~\cite{zelikman2022star,ho2023large,hsieh2023distilling,magister2023teaching}.
We generate rationales only for supported and partially supported labels, so the
teacher is never asked to justify diagnoses the evidence does not substantiate.
The layer lets us study a central trade-off for trustworthy deployment: whether
equipping a model to emit evidence-grounded rationales aids or hinders its
diagnostic accuracy.

\section{Modality Contribution}
\label{app:modality}
\begin{figure}[t]
  \centering
  \includegraphics[width=\columnwidth]{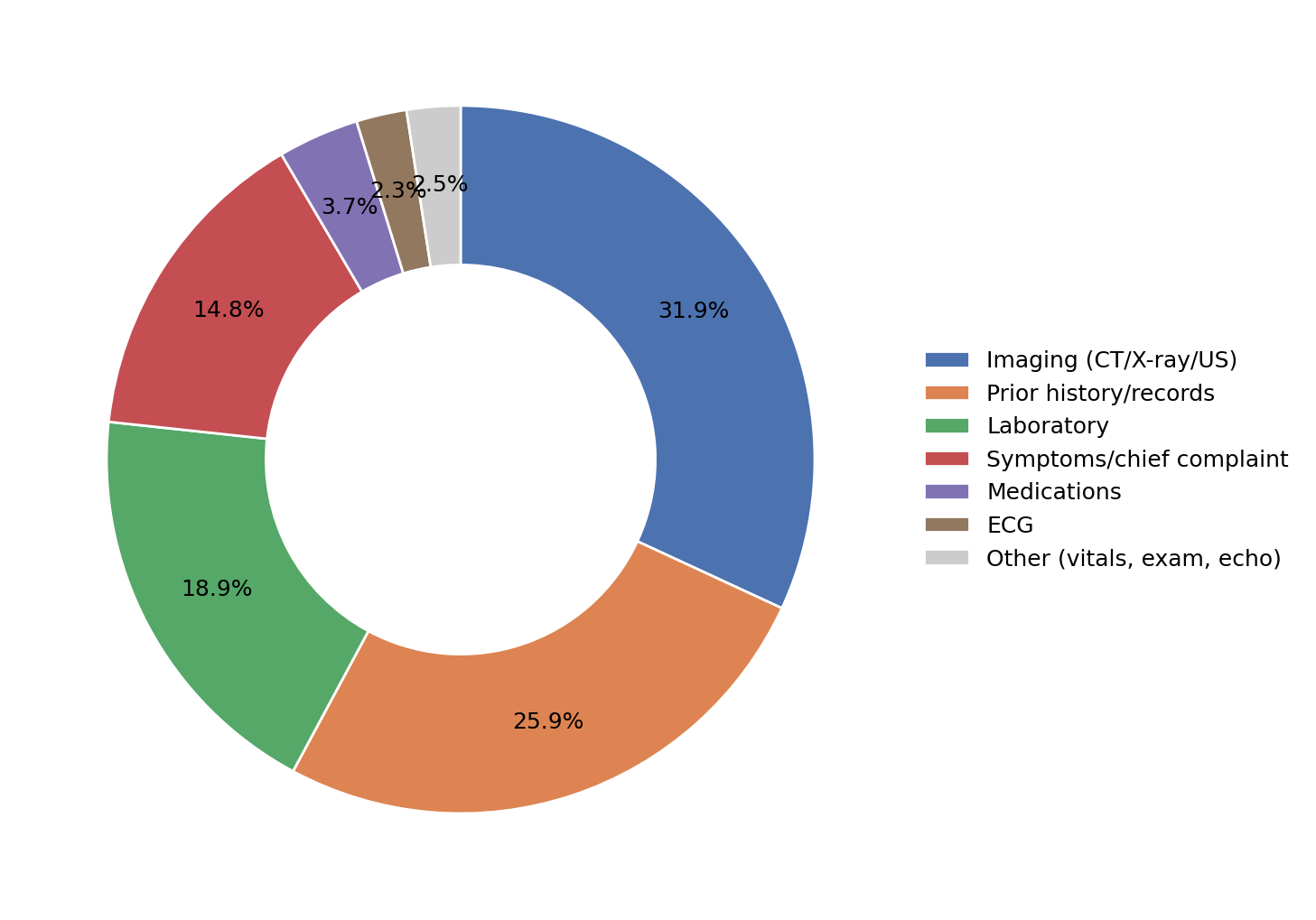}
  \caption{Exploratory primary supporting-evidence modality across all diagnoses,
  approximated from the generated auxiliary rationales. Direct test evidence
  (imaging and laboratory) accounts for ${\sim}51\%$, prior history and records
  for $26\%$.}
  \label{fig:modality}
\end{figure}

As an exploratory analysis, we use the generated rationales to approximate which
evidence source each retained diagnosis primarily relies on, classifying the
reasoning into one of nine evidence types with the judge model
(Figure~\ref{fig:modality}). Imaging (CT/X-ray/ultrasound) is the largest single
source ($31.9\%$), followed by laboratory results ($18.9\%$), so direct
early-time test evidence underlies roughly half of all diagnoses. Prior history
and records form the second largest source ($25.9\%$) and symptoms and chief
complaint $14.8\%$, while physiological signals contribute only marginally (ECG
$2.3\%$, echocardiography $0.3\%$), reflecting the relative prevalence of
waveform-decidable conditions in the ED. The breakdown mirrors the
supported--partial gap of Section~\ref{sec:direct}: diagnoses grounded in direct
test evidence coincide with the reliably predictable supported labels, whereas
those resting on prior history align with the harder, partially supported ones.

\begin{figure}[t]
  \centering
  \includegraphics[width=\columnwidth]{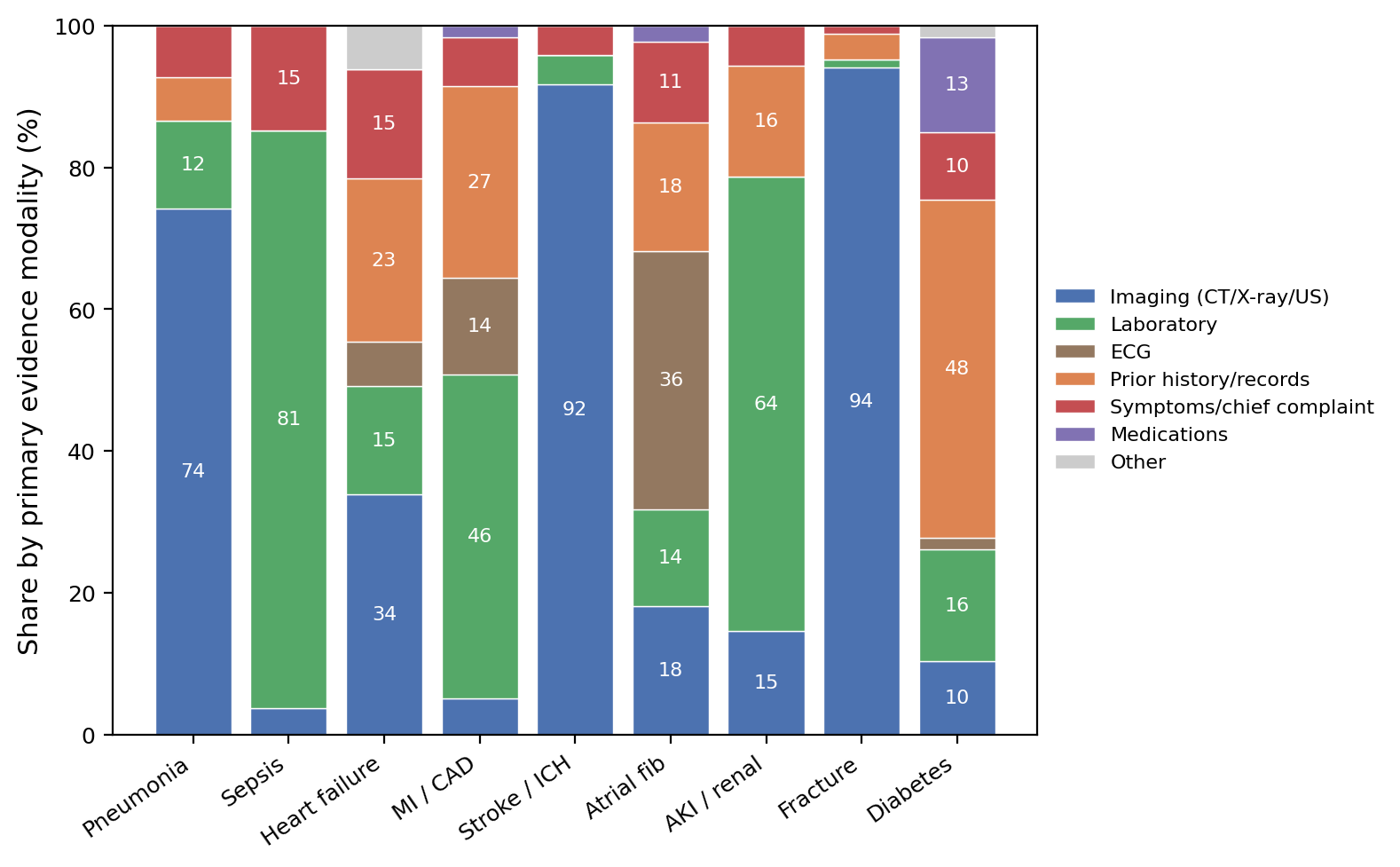}
  \caption{Exploratory primary supporting-evidence modality for sample diseases,
  approximated from the generated auxiliary rationales. The decisive modality
  shifts sharply from one condition to another.}
  \label{fig:disease_modality}
\end{figure}

The critical modality is strongly disease-specific
(Figure~\ref{fig:disease_modality}). Imaging is decisive for structural and
traumatic conditions, underlying $94\%$ of fracture and $92\%$ of
stroke/intracranial-hemorrhage diagnoses; laboratory results dominate metabolic
and infectious ones, at $81\%$ of sepsis and $64\%$ of acute kidney injury.
Electrocardiography, a minor contributor overall, is the most decisive single
modality for atrial fibrillation ($36\%$) and a key adjunct for myocardial
infarction ($14\%$, alongside $46\%$ laboratory). Chronic conditions such as
diabetes and heart failure instead rely on prior history and records ($48\%$ and
$23\%$). These patterns are clinically coherent and suggest that no single
modality suffices for admission diagnosis.

\section{Example Record}
\label{app:example}
A representative encounter (subarachnoid hemorrhage). The serialized input aggregates all
admission-time modalities (absent fields marked \texttt{None}); the assistant turn holds the
chain-of-thought and the free-text diagnosis; \texttt{kept\_verdicts} records the evidence
category. Long lines are truncated for display.

\begin{lstlisting}[basicstyle=\ttfamily\scriptsize,breaklines=true,frame=single]
# INPUT (user turn)
Demographics: M, 60yo, WHITE. From EMERGENCY ROOM, arrival WALK IN.
Chief complaint: SAH, Transfer
Triage: T97.8 HR52.0 RR16.0 SpO297.0 BP162.0/90.0 acuity2.0
ED serial vitals: HR52.0 SpO297.0 | HR61.0 SpO297.0
Home meds: None
Baseline (OMR): BP None, Wt None, Ht None, BMI None
Baseline labs (prior): creatinine 1.0, hemoglobin 15.2
Past medical history (prior dx): None
Past ED diagnoses: None
Past medical history (from prior discharge note): None
ECG: Sinus bradycardia.; Inferior T wave changes are nonspecific;
     Borderline ECG | HR54 PR180 QRS108 QT478ms
Echocardiogram: None
Radiology (3): [CTA HEAD W AND W/O C] F: CT HEAD WITHOUT CONTRAST: focal
     hyperdensity within the suprasellar, interpeduncular, and prepontine
     cisterns, consistent with subarachnoid hemorrhage ...
Initial labs (26): White Blood Cells 11.7*; Glucose 143*; RBC 4.45*; ...

# ASSISTANT (target)
<think> The patient presents with a confirmed subarachnoid hemorrhage based
on CT imaging: the non-contrast CT head demonstrates focal hyperdensity
within the suprasellar, interpeduncular, and prepontine cisterns, consistent
with SAH. The patient is hemodynamically stable with sinus bradycardia on
ECG and unremarkable labs. </think>
<answer>SUBARACHNOID HEMORRHAGE</answer>

# kept_verdicts
[{"dx": "SUBARACHNOID HEMORRHAGE", "verdict": "supported"}]
\end{lstlisting}

\section{Prompts}
\label{app:prompts}

We use four prompts in the pipeline: an evidence verifier (label verification), a teacher for
chain-of-thought generation, a zero-shot inference prompt, and an LLM-as-judge for evaluation.
Placeholders \texttt{\{INPUT\}}, \texttt{\{DX\}}, \texttt{\{G\}}, \texttt{\{P\}} are filled per
encounter.

\newtcolorbox{promptbox}[1]{colback=gray!4,colframe=gray!55!black,fonttitle=\bfseries,
  title=#1,boxrule=0.5pt,left=3pt,right=3pt,top=2pt,bottom=2pt,
  before skip=6pt,after skip=6pt}

\begin{promptbox}{B.1\quad Evidence Verifier}
\small\ttfamily
You are a clinical evidence auditor for an admission-time diagnosis dataset. Given the admission-anchored early-workup INPUT and the RECORDED DIAGNOSES, judge for EACH diagnosis whether the INPUT contains evidence to support it within the specified evidence window.\\[2pt]
Verdict options:\\
$\bullet$ \textbf{supported}: clear/direct evidence in input (labs, imaging, vitals, exam, or
clearly consistent presentation)\\
$\bullet$ \textbf{partial}: only indirect/suggestive, or supported only by past history / home
medications (chronic comorbidity), not acute presentation\\
$\bullet$ \textbf{unsupported}: no evidence, or input clearly points to a different problem\\[2pt]
INPUT: \{INPUT\}\\
RECORDED DIAGNOSES: \{DXS\}\\
Output ONLY JSON: \{"verdicts":[\{"dx":...,
"verdict":...,"reason":<=15 words\}]\}
\end{promptbox}

\begin{promptbox}{B.2\quad Gold-Conditioned Rationale Generation (teacher)}
\small\ttfamily
You are an expert emergency physician reasoning through a patient at admission. FIRST work
through the relevant findings (vitals, labs with values, imaging, history, medications),
interpret them, then arrive at each diagnosis as a CONCLUSION at the end.\\[2pt]
Rules: do not start a sentence with the diagnosis name; use ONLY findings present in the input;
if a diagnosis is a known chronic condition, say so; if imaging/labs are negative or equivocal,
state the evidence is limited and the diagnosis is presumptive; be succinct.\\[2pt]
INPUT: \{INPUT\}\\
CONFIRMED DIAGNOSES (reasoning must converge to these): \{DX\}\\
Output: \texttt{<think>}[evidence-first reasoning]\texttt{</think>}
\texttt{<answer>}\{DX\}\texttt{</answer>}
\end{promptbox}

\begin{promptbox}{B.3\quad Zero-shot Inference}
\small\ttfamily
You are an expert emergency physician. Given ONLY the admission-anchored early-workup presentation, reason from
the evidence and determine THIS encounter's diagnoses. You are NOT given the answer.\\[2pt]
INPUT: \{INPUT\}\\
Output: \texttt{<think>}[brief reasoning]\texttt{</think>}
\texttt{<answer>}diagnosis1; diagnosis2\texttt{</answer>}
\end{promptbox}

\begin{promptbox}{B.4\quad LLM-as-Judge}
\small\ttfamily
You are a clinical coding judge. GOLD diagnoses (truth) and PRED diagnoses (model) for one
admission. Two diagnoses MATCH if they refer to the same clinical condition
(synonyms/abbreviations/specificity differences count as a match). Compute a 1-to-1 matching
between GOLD and PRED (each item used at most once).\\[2pt]
GOLD: \{G\}\quad PRED: \{P\}\\
Return ONLY JSON: \{"matched\_pairs": <number>\}
\end{promptbox}

\section{Training and Inference Configuration}
\label{app:config}
\begin{figure}[t]\centering
  \includegraphics[width=0.8\columnwidth]{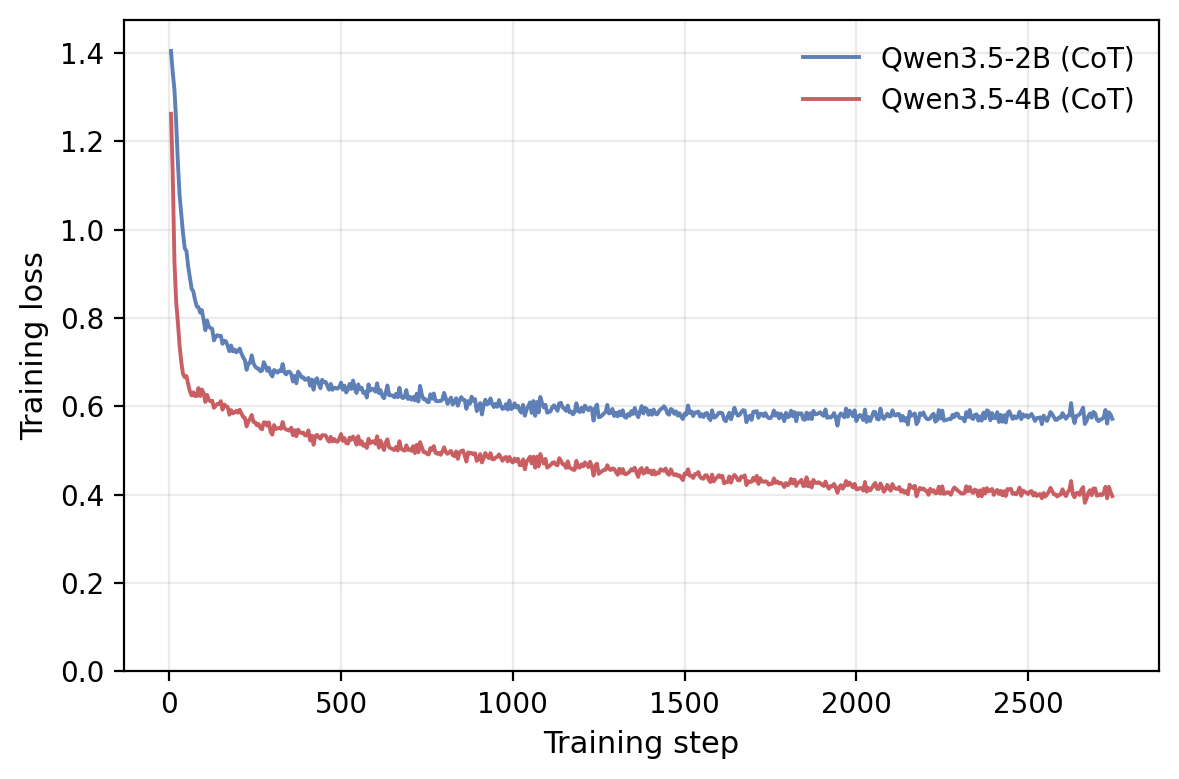}
  \caption{Training loss for the 2B and 4B rational-SFT models (completion-only loss, one epoch,
  effective batch size 48). Both converge smoothly; the 4B model reaches a lower loss
  (0.40 vs.\ 0.57), consistent with its higher capacity.}
  \label{fig:train}
\end{figure}

All models are fully fine-tuned for one epoch with a completion-only objective
(prompt tokens masked): AdamW, peak learning rate $1\times10^{-5}$, cosine
schedule, $3\%$ warmup, effective batch size $48$, maximum sequence length
$3072$, \texttt{bfloat16}, and gradient checkpointing. The 2B models use
distributed data parallelism over three GPUs; the 4B models use DeepSpeed ZeRO-2
with optimizer CPU offload. Inference uses greedy decoding with up to $2048$ new
tokens; teacher CoT traces are generated at temperature $0.3$, and the verifier
and judge run at temperature $0$. Both model sizes converge within the single
epoch (Figure~\ref{fig:train}), the 4B model reaching a lower final loss
($0.40$ vs.\ $0.57$).

\section{Evidence Window and the Cost of Rationales}
\label{app:evidence_window}
To assess sensitivity to evidence accrued after the admission anchor, we
re-evaluate the post-trained models against the same evidence-supported gold set
at three cutoffs $W\in\{0,6,24\}$\,h (Table~\ref{tab:window}). Performance rises
monotonically with $W$ (4B-CoT micro-F1 $0.434\!\to\!0.463\!\to\!0.476$),
reflecting the gradual accumulation of the early workup. Crucially, even at
$W{=}0$---where the input contains only records time-stamped no later than
hospital admission---the fine-tuned model reaches $0.434$, well above the
strongest zero-shot baseline ($0.36$). Our central finding, that a small
task-aligned model substantially outperforms much larger zero-shot LLMs, is
therefore not an artifact of the chosen window.

The two supervision formats expose a trade-off between auditability and accuracy.
A faithful rationale is a harder learning target than the answer alone, and the
direct format matches or exceeds CoT at every window. The cost is one of
\emph{learnability} rather than a limitation of reasoning: the CoT--direct F1 gap
contracts sharply with capacity, from $-0.046$ at 2B to $-0.008$ at 4B
($W{=}6$). A sufficiently capable model thus recovers nearly all of the accuracy
lost to producing a rationale while retaining the rationale itself, suggesting
that interpretability of this kind is paid for by scale.

\begin{table}[t]
  \centering
  \small
  \setlength{\tabcolsep}{5pt}
  \caption{Window-sensitivity of all post-trained models, scored against the same
  evidence-supported (supported$+$partial) gold set. The 4B models dominate their
  2B counterparts; the 2B-CoT model responds non-monotonically, reflecting the
  instability of long generated rationales at small scale.}
  \label{tab:window}
  \begin{tabular}{llccc}
    \toprule
    Model & Window & micro-P & micro-R & micro-F1 \\
    \midrule
    \multirow{3}{*}{2B-CoT}
      & $W{=}0$ (\(\le\)admit) & 0.331 & 0.253 & 0.287 \\
      & $W{=}6$h               & 0.445 & 0.380 & 0.410 \\
      & $W{=}24$h              & 0.370 & 0.307 & 0.335 \\
    \midrule
    \multirow{3}{*}{2B-direct}
      & $W{=}0$ (\(\le\)admit) & 0.469 & 0.404 & 0.434 \\
      & $W{=}6$h               & 0.496 & 0.422 & 0.456 \\
      & $W{=}24$h              & 0.556 & 0.451 & 0.498 \\
    \midrule
    \multirow{3}{*}{4B-CoT}
      & $W{=}0$ (\(\le\)admit) & 0.485 & 0.393 & 0.434 \\
      & $W{=}6$h               & 0.502 & 0.429 & 0.463 \\
      & $W{=}24$h              & 0.523 & 0.437 & 0.476 \\
    \midrule
    \multirow{3}{*}{4B-direct}
      & $W{=}0$ (\(\le\)admit) & 0.496 & 0.404 & 0.446 \\
      & $W{=}6$h               & 0.529 & 0.424 & 0.471 \\
      & $W{=}24$h              & 0.573 & 0.447 & 0.502 \\
    \bottomrule
  \end{tabular}
\end{table}

\section{Chain-of-Thought Rationale Examples}
\label{app:cot-examples}
We provide qualitative examples of the auxiliary gold-conditioned rationales
introduced in Section~\ref{sec:rationale}. These are gold-conditioned post-hoc
explanations used only as an exploratory training signal---not validated clinical
reasoning traces, and not part of the benchmark evaluation target.

Table~\ref{tab:cot-examples} shows a representative rationale at each quality
level as rated by reviewer consensus: a \emph{correct} rationale cites a specific
early-time finding that substantiates the diagnosis; a \emph{partial} one reaches
the right label but leans on prior history without explaining the acute
presentation; an \emph{incorrect} one asserts a conclusion not grounded in the
input.

Figure~\ref{fig:cot_examples} contrasts a supported and a partially supported
diagnosis in full. The supported case (subarachnoid hemorrhage) is grounded in
direct early-time evidence---a CT finding the rationale points to
explicitly---whereas the partially supported case (thyroid cancer) is inferable
only from indirect cues such as prior history, illustrating the performance gap
between the two categories in Table~\ref{tab:full}.

Figure~\ref{fig:human_vs_llm_cot} contrasts a clinician's rationale with our
model's chain-of-thought on the same case. Both reach the correct diagnosis, but
the clinician reasons forward in far fewer words---comparing against baseline,
raising a competing concern, recommending the next test---whereas the generated
rationale is a confident post-hoc justification.

\begin{table*}[t]
  \centering
  \small
  \renewcommand{\arraystretch}{1.25}
  \caption{Representative (gold-conditioned) auxiliary rationales at each
  quality level, with the consensus reviewer assessment. \emph{Correct}
  rationales cite a specific early-time finding; \emph{partial} ones reach
  the right label but lean on history without explaining the acute
  presentation; \emph{incorrect} ones assert a conclusion not grounded in the
  input.}
  \label{tab:cot-examples}
  \begin{tabularx}{\textwidth}{@{}l l X@{}}
    \toprule
    \textbf{Level} & \textbf{Diagnosis} & \textbf{Rationale excerpt and reviewer assessment} \\
    \midrule
    \textsc{Correct}
      & Dialysis-catheter infection
      & ``Chief complaint \emph{infection dialysis catheter}; prior catheter
        infections; triage hypotension (BP 80/40) acutely worse than baseline
        (88/50) with persistent tachycardia\ldots'' \newline
        \textit{Assessment:} the conclusion is anchored to specific present
        findings (catheter site, hypotension, sepsis physiology). Unanimously
        accepted. \\
    \addlinespace[2pt]
    \textsc{Partial}
      & Thyroid carcinoma
      & ``45-year-old with dysphagia following radiation; history of thyroid
        malignancy\ldots the presentation strongly suggests a treatment
        complication such as radiation esophagitis\ldots'' \newline
        \textit{Assessment:} the label (a \emph{known} condition) is correct, but
        the reasoning restates history and does not explain the \emph{acute}
        complaint. Majority rated partial. \\
    \addlinespace[2pt]
    \textsc{Incorrect}
      & CHF; anemia
      & ``74-year-old with dyspnea, COPD (FEV1 42\%), transient desaturation\ldots
        elevated BP\ldots'' \newline
        \textit{Assessment:} the rationale develops a COPD/respiratory picture
        but offers no admission-time evidence for heart failure or anemia (no
        BNP, echo, or hemoglobin cited). Unanimously rated incorrect. \\
    \bottomrule
  \end{tabularx}
\end{table*}

\begin{figure*}[t]\centering
\begin{tcolorbox}[colback=green!4,colframe=green!45!black,title=\textbf{Supported example}: 
Subarachnoid Hemorrhage,width=0.48\textwidth,fonttitle=\bfseries,nobeforeafter,
before upper={\parindent0pt}]
\textbf{Chief complaint:} SAH, Transfer\\[2pt]
\texttt{<think>} The patient presents with a confirmed subarachnoid hemorrhage 
\underline{based on CT imaging}. The non-contrast CT head demonstrates focal hyperdensity 
within the suprasellar, interpeduncular, and prepontine cisterns, \underline{consistent with 
subarachnoid hemorrhage} \ldots \texttt{</think>}\\[2pt]
\texttt{<answer>} SUBARACHNOID HEMORRHAGE \texttt{</answer>}\\[4pt]
{\footnotesize\itshape Diagnosis traced to direct admission-time evidence (CT).}
\end{tcolorbox}
\hfill
\begin{tcolorbox}[colback=orange!5,colframe=orange!55!black,title=\textbf{Partial example}: 
Thyroid Cancer,width=0.48\textwidth,fonttitle=\bfseries,nobeforeafter,
before upper={\parindent0pt}]
\textbf{Chief complaint:} Dysphagia, s/p radiation\\[2pt]
\texttt{<think>} A 45-year-old male presenting with dysphagia following radiation therapy. 
His \underline{past medical history} includes malignant neoplasm of the thyroid. The 
diagnosis CA THYROID refers to his \underline{known history} of thyroid cancer; the 
presentation is likely a sequela of prior treatment \ldots \texttt{</think>}\\[2pt]
\texttt{<answer>} CA THYROID \texttt{</answer>}\\[4pt]
{\footnotesize\itshape Diagnosis inferred from prior history, not a current finding.}
\end{tcolorbox}
\caption{Chain-of-thought traces for a \textbf{supported} (left) and a \textbf{partially 
supported} (right) diagnosis. Supported diagnoses are grounded in direct early-time 
evidence (e.g., CT findings), whereas partially supported diagnoses are inferable only from 
indirect cues such as prior history---explaining the large performance gap between the two 
categories (Table~\ref{tab:full}).}
\label{fig:cot_examples}
\end{figure*}

\begin{figure*}[t]\centering
\begin{minipage}[t]{0.48\textwidth}
\begin{tcolorbox}[colback=violet!4,colframe=violet!55!black,
  title={\textbf{Human clinician} --- 70 words},
  fonttitle=\bfseries, width=\linewidth]
\small
Middle-aged woman with asthma and COPD. On arrival the respiratory rate is
markedly elevated with falling SpO$_2$, and BP is \underline{clearly above her
baseline}; the venous gas shows a high pCO$_2$. I favour an
\underline{asthma exacerbation / AECOPD}. The ECG changes, however, make
\underline{myocardial ischaemia likely}---I would \underline{obtain cardiac
enzymes} next.

\smallskip
{\footnotesize\itshape Forward, uncertainty-aware: commits to a diagnosis, raises
a competing concern, and names the next test.}
\end{tcolorbox}
\end{minipage}%
\hfill
\begin{minipage}[t]{0.48\textwidth}
\begin{tcolorbox}[colback=teal!4,colframe=teal!55!black,
  title={\textbf{Qwen3.5-4B (CoT)} --- 200 words},
  fonttitle=\bfseries, width=\linewidth]
\small
51-year-old with documented chronic obstructive asthma; home albuterol and
budesonide--formoterol. On arrival tachypnoeic (RR 24), tachycardic (HR 125--130),
hypoxic (SpO$_2$ 93\%) improving with treatment. Chest X-ray shows no acute
process (rules out pneumonia/pneumothorax); the blood gas shows respiratory
acidosis (pH 7.27, pCO$_2$ 51). The picture aligns with an
\underline{acute exacerbation of chronic obstructive asthma}.

\smallskip
{\footnotesize\itshape Systematic and correctly calibrated to a single diagnosis,
but a \emph{post-hoc justification}: no uncertainty, competing differential, or
next-action.}
\end{tcolorbox}
\end{minipage}

\caption{Human vs.\ our fine-tuned model's chain-of-thought for the same admission
(asthma/COPD exacerbation). Both reach the correct diagnosis, and the fine-tuned
model is well calibrated---committing to a single diagnosis rather than
over-listing. Yet the clinician additionally reasons \emph{forward}: comparing
against baseline, raising a competing concern (ischaemia), and recommending the
next test, whereas the generated rationale is a confident post-hoc justification
of one label.}
\label{fig:human_vs_llm_cot}
\end{figure*}

\section{Clinical Review of Chain-of-Thought Rationales}
\label{app:cot_review}
\begin{figure}[h]\centering
  \begin{subfigure}[b]{\columnwidth}\centering
    \includegraphics[width=\linewidth]{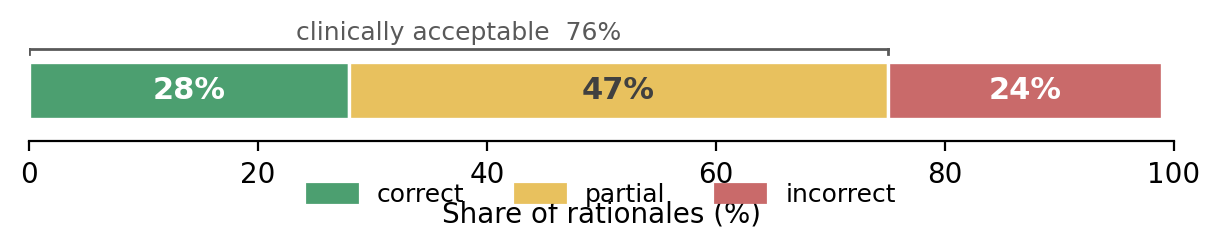}
    \caption{Clinical verdict on the $1000$ reviewed rationales.}
    \label{fig:cot_verdict}
  \end{subfigure}
  \\[4pt]
  \begin{subfigure}[b]{\columnwidth}\centering
    \includegraphics[width=\linewidth]{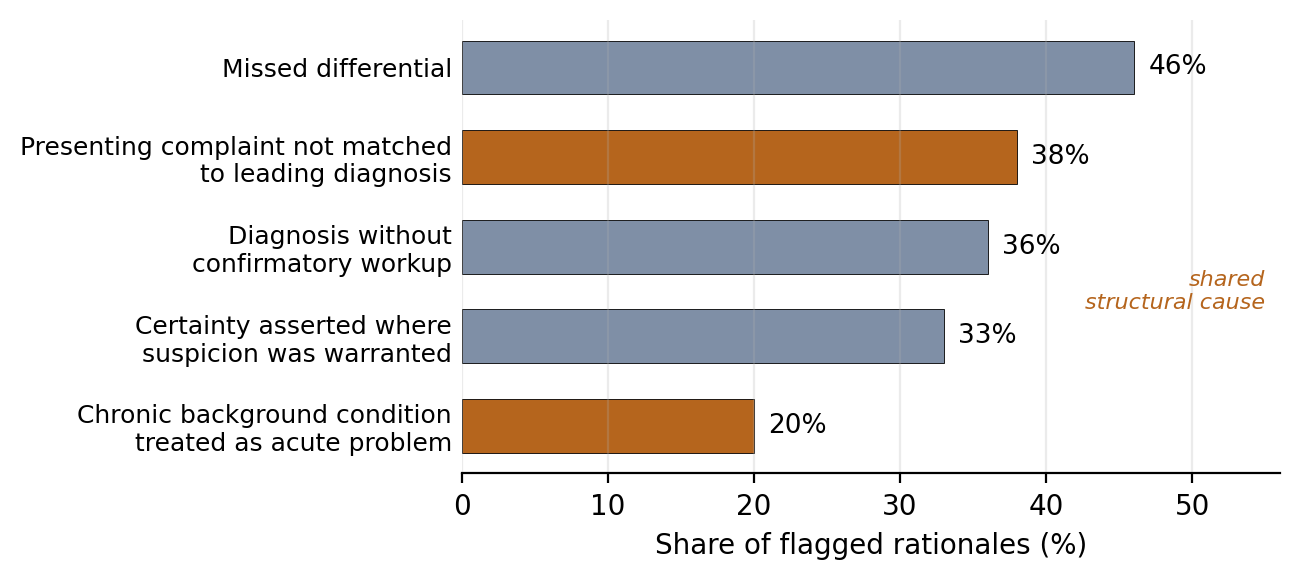}
    \caption{Faults across the flagged rationales; categories are not
    mutually exclusive and the two highlighted ones share a common cause.}
    \label{fig:cot_faults}
  \end{subfigure}
  \caption{Clinical review of gold-conditioned rationales.}
  \label{fig:cot_review}
  \Description{Two panels. Top: a stacked bar showing 28 percent correct,
  47 percent partial, 24 percent incorrect. Bottom: a horizontal bar chart of five
  fault categories ranging from 20 to 46 percent, two of them highlighted.}
\end{figure}

A board-certified clinician reviewed the auxiliary rationales for $1{,}000$ randomly
sampled test encounters, presented with the admission-time input and the reference
diagnosis. Because the rationales are gold-conditioned, the final diagnosis is
correct by construction; the reviewer therefore assessed only whether the
reasoning toward it was clinically sound and grounded in evidence present in the
input, recording the specific fault otherwise. $76\%$ were judged clinically
acceptable, though fewer than a third fully sound
(Figure~\ref{fig:cot_review}a), and the faults recorded across the flagged
rationales concentrate on a few recurring patterns (Figure~\ref{fig:cot_review}b).

\paragraph{Background conditions displace the acute problem.} A single mechanism
underlies two of these faults: chronic conditions recur throughout the prior
history, and the model promotes them to the leading diagnosis in place of the
acute event that prompted the visit. For a patient with cirrhosis presenting with
altered mental status, the encounter concerns hepatic encephalopathy---cirrhosis
is the standing background against which that event occurs---yet the model
returns cirrhosis as the primary, at times sole, diagnosis. The pattern recurs
where a patient with COPD develops pneumonia and the model reports COPD rather
than pneumonia with an acute exacerbation.

The failure is one of assignment rather than of knowledge: the model states the
pathophysiology correctly but misidentifies which condition the encounter is
about. Prior history enters the input as undifferentiated text, so a condition
mentioned often acquires salience proportional to its frequency rather than its
acuity. The remedy is therefore representational rather than a matter of
prompting: prior history should be demarcated as a distinct field, as demographics
already are, and consumed as risk context rather than as candidate diagnoses, so
that the model reasons over what is \emph{new} in the encounter conditioned on
what is chronically true of the patient. We regard this as the review's most
actionable finding, since it identifies a defect that improved rationale
supervision cannot correct and that generalizes to any admission-time diagnosis
system built over longitudinal records.

\section{Human Review Protocol}
\label{app:review}
Four expert reviewers independently rated $1{,}000$ randomly sampled training
instances along two axes. For each instance the reviewer was shown the
admission-time input, the generated chain-of-thought (CoT), the predicted
diagnoses, and the assigned evidence verdict.

\paragraph{Axis 1: CoT reasoning quality.} Whether the reasoning is clinically
sound and faithful to the input:
\begin{itemize}\itemsep1pt
\item \textbf{Correct} --- clinically valid reasoning in which every cited finding
is actually present in the input and plausibly supports the stated diagnosis.
\item \textbf{Partial} --- largely sound but with a minor flaw, e.g.\ overstated
certainty, a weak evidence link, or a small misreading of a value.
\item \textbf{Incorrect} --- the reasoning hallucinates or contradicts the input,
or the conclusion does not follow from the cited evidence.
\end{itemize}

\paragraph{Axis 2: Evidence-class assignment.} Whether the verifier's label
(supported / partially supported / unsupported) matches the evidence actually
available at admission:
\begin{itemize}\itemsep1pt
\item \textbf{Appropriate} --- the class correctly reflects the evidence, e.g.\
\emph{supported} when a direct finding confirms the diagnosis, \emph{partial} when
only history or home medications are suggestive.
\item \textbf{Borderline} --- defensible but on the supported/partial margin
(direct vs.\ indirect evidence), where reasonable raters could differ.
\item \textbf{Wrong} --- a clear mismatch, e.g.\ \emph{supported} assigned with no
admission-time finding, or \emph{partial} despite a confirmatory image.
\end{itemize}

\section{Input-Modality Ablation}
\label{app:ablation}
To test whether each diagnosis \emph{causally} depends on the attributed modality,
we remove one modality at a time from the input at inference (laboratory results,
radiology, ECG, prior history, or echocardiography) and re-evaluate the 4B-CoT
model on a $2{,}000$-encounter subset, measuring the per-disease change in F1
(Figure~\ref{fig:ablate_heat}). The effects are clinically coherent and mirror the
attribution of Appendix~\ref{app:modality}. Removing radiology collapses
imaging-confirmed conditions---fracture ($-0.37$ F1), stroke / intracranial
hemorrhage ($-0.24$), pneumonia ($-0.22$); removing laboratory results most
degrades acute kidney injury ($-0.16$) and myocardial infarction ($-0.08$), which
depend on creatinine and troponin; and ECG is the only modality whose removal
hurts atrial fibrillation ($-0.06$), the canonical ECG diagnosis. In aggregate,
radiology is the most load-bearing modality ($-0.09$ overall), followed by
laboratory results ($-0.05$) and prior history ($-0.03$), whereas ECG and
echocardiography have little average effect, consistent with their narrow
indications and lower coverage (Figure~\ref{fig:coverage}). No single modality
therefore suffices: admission diagnosis requires integrating evidence whose
relevance shifts from one disease to another. Two caveats apply: small disease
buckets (e.g.\ sepsis, $n{=}15$) give noisy estimates, and an inference-time
ablation measures a fixed model's reliance on each modality rather than its
importance under retraining.

\begin{figure}[t]\centering
  \includegraphics[width=\columnwidth]{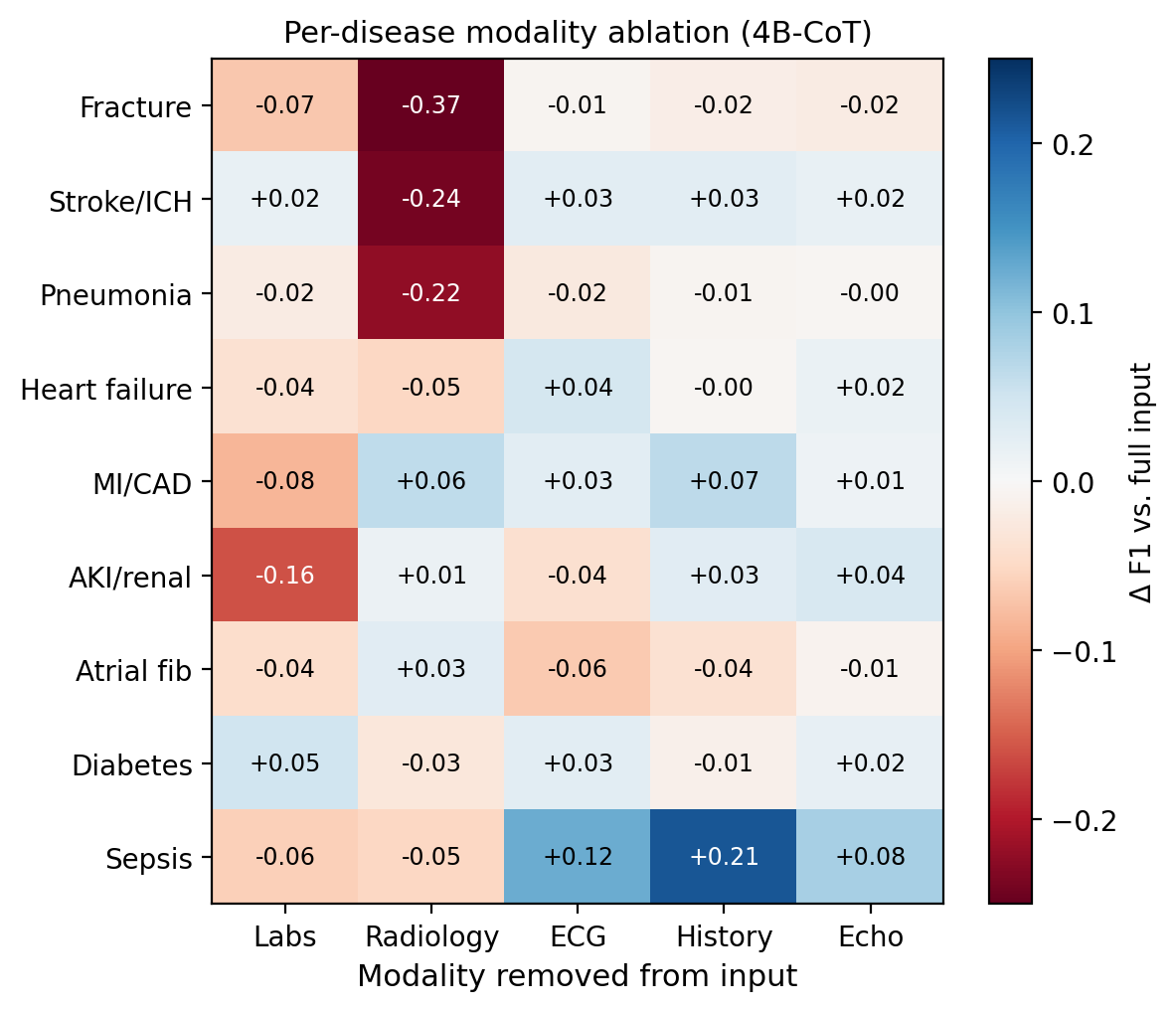}
  \caption{Per-disease input-modality ablation (4B-CoT) on a $2{,}000$-encounter
  subset: change in F1 when each modality is removed at inference. Red indicates a
  performance drop. Estimates for low-frequency conditions are unstable and should
  be read with caution.}
  \Description{Heatmap of per-disease F1 change when input modalities are removed.
  Rows are diseases, columns are ablated modalities. Cells are red where removal
  degrades performance and blue where it does not.}
  \label{fig:ablate_heat}
\end{figure}

\section{Data and Code Availability}
\label{app:availability}
\textbf{Code.} The complete construction and evaluation pipeline (cohort
extraction, input serialization, label filtering, evidence verification,
chain-of-thought generation, training, inference, and LLM-judge evaluation),
together with all prompts, the fixed patient-level split, and the cached verifier
and judge outputs, is available at \url{https://github.com/jimmylihui/EarlyDx}.
The repository contains code only and no patient data.

\textbf{Dataset and models.} EarlyDx derives from MIMIC-IV, distributed under the
PhysioNet Credentialed Health Data License. In accordance with that Data Use
Agreement we do \emph{not} release the dataset or the fine-tuned weights publicly;
both are made available through PhysioNet to researchers holding MIMIC-IV
credentials, inheriting the access restrictions of the parent database.

\textbf{Reproduction.} Credentialed users can regenerate EarlyDx end-to-end from
the released code given local access to the MIMIC-IV, MIMIC-IV-ED, and
MIMIC-IV-Note modules; the fixed split and cached verifier/judge outputs make the
benchmark and all reported numbers exactly reproducible. Identifiers and access
dates of all evaluated API models, and versions of the open-source models, are
listed in the repository. MIMIC-IV itself is available at
\url{https://physionet.org/content/mimiciv/}.

\section{Ethical Considerations}
\noindent\textbf{Data handling.} All components that process the full corpus---the
evidence verifier, the teacher used for rationale generation, and the semantic
judge---are open-weight models run on local infrastructure; no MIMIC-derived text
leaves our institutional environment for these stages. The two proprietary
zero-shot systems were evaluated on the $6{,}975$ test encounters through Azure
OpenAI with human review disabled (GPT-5.5) and the Anthropic API under a
zero-retention agreement (Claude Opus 4.8), deployments for which we verified that
inputs are not retained, not used for training, and not subject to routine human
review, as the PhysioNet credentialed data use agreement requires. Remaining
open-weight baselines were likewise run locally, and the released pipeline is
provider-agnostic: because verifier and judge outputs are cached, reproducing our
numbers requires no further queries to any external service.

\noindent\textbf{Provenance and scope.} EarlyDx derives solely from MIMIC-IV,
which is HIPAA-deidentified and released under IRB approval; we collected no new
data and screened all generated text for residual identifiers. Per the MIMIC data
use agreement, the dataset and fine-tuned weights are shared only with
credentialed users through PhysioNet, while the data-free code is public. EarlyDx
is a research benchmark, not a clinical tool: it is not validated for patient
care, and any downstream use would require prospective validation and human
oversight.

\section{Datasheet for EarlyDx}
\label{app:datasheet}
Table~\ref{tab:datasheet} gives a datasheet for EarlyDx covering motivation,
composition, collection, preprocessing, intended and out-of-scope uses,
distribution terms, versioning, and maintenance. Two entries warrant emphasis: the
auxiliary rationale layer is gold-conditioned and must not be repurposed as
reasoning supervision, and the datasheet records which components are frozen at
release and which must remain executable to score future submissions.

\begin{table*}[t]\centering\small
\caption{Datasheet for EarlyDx, following Gebru et al.}
\label{tab:datasheet}
\renewcommand{\arraystretch}{1.25}
\begin{tabular}{@{}p{0.16\textwidth}p{0.80\textwidth}@{}}
\toprule
\textbf{Motivation} & Created to evaluate open-ended diagnosis generation from
evidence available at hospital admission. Existing benchmarks are unsuited to this
setting: they use closed code sets and supervise with discharge diagnoses coded
over the full inpatient stay. Created by the authors; no external body directed
its design. \\
\midrule
\textbf{Composition} & One instance per ED encounter resulting in admission:
a serialized text prompt of the admission-time record, plus free-text
ED-encounter diagnoses each annotated with an evidence class (supported /
partially supported / unsupported). Derived from MIMIC-IV~v3.1, MIMIC-IV-ED~v2.2,
MIMIC-IV-Note~v2.2, MIMIC-IV-ECG~v1.0; contains no data external to MIMIC-IV.
Cohort composition, modality coverage, and subgroup breakdowns. \\
\midrule
\textbf{Collection} & Programmatic selection of every ED stay with a linked
hospital admission and at least one substantive ED diagnosis, excluding symptom
and ill-defined codes. No sampling: the cohort is the
complete set satisfying these criteria. Source records are de-identified by the
MIMIC-IV providers; generated text is additionally screened for residual
identifiers. \\
\midrule
\textbf{Preprocessing} & Records clipped to a configurable window around admission
time and serialized in fixed field order; absent fields marked \texttt{None}.
Evidence classes assigned by an LLM verifier; auxiliary rationales generated
conditioned on the reference diagnosis. Raw MIMIC-IV tables
remain available from PhysioNet, so preprocessing is reproducible from source. \\
\midrule
\textbf{Intended uses} & Benchmarking open-ended diagnosis generation from
admission-time evidence; studying label noise and temporal leakage in
MIMIC-derived prediction tasks; auditing models trained on discharge-coded
labels. \\
\midrule
\textbf{Uses out of scope} & \textbf{Not for clinical decision support,
deployment, or training any model for clinical use.} Not validated for patient
care; single-institution; references are administrative billing codes, not
adjudicated clinical diagnoses. Cannot be used to study the admit-versus-discharge
decision: discharged encounters are excluded by construction. Two filters shape
the population---no ED-discharged encounters, and no encounters whose diagnoses
are all unsupported---biasing the retained set toward conditions with visible
admission-time evidence. \\
\midrule
\textbf{Rationale layer} & The released chain-of-thought rationales are
\emph{gold-conditioned}: generated with the reference diagnosis in context, they
are post-hoc justifications rather than the reasoning that produced the
diagnosis, and multi-reviewer audit accepted only a minority as fully clinically
sound. They must not be used as reasoning supervision, as
clinical explanations, or as evidence of a model's reasoning process. This
statement is reproduced in the repository README and in per-file metadata. \\
\midrule
\textbf{Distribution} & Released through PhysioNet under credentialed access
governed by the MIMIC-IV data use agreement. All derived layers---evidence
classes, rationales, splits, and cached judge decisions---are distributed under
the same credentialed-access terms as the source records and may not be
redistributed to non-credentialed parties. Construction and evaluation code, the
judge configuration, and documentation contain no MIMIC-derived text and are
released publicly. Model-generated annotations are redistributed subject to the
generating providers' terms. \\
\midrule
\textbf{Versioning} & \texttt{MAJOR.MINOR.PATCH}: \textsc{major} for a change in
cohort or label semantics, \textsc{minor} for added layers or splits,
\textsc{patch} for corrections. Superseded versions remain retrievable, so a
future citation to \texttt{EarlyDx v1.2} resolves to a fixed artifact.
Machine-readable metadata is provided in Croissant format. \\
\midrule
\textbf{Maintenance} & Maintained by Jiahui Li (\texttt{jl57095@uga.edu}) for at
least three years following release. Errata are reported via the repository issue
tracker; confirmed errors are corrected in a new patch version with a changelog
entry. \\
\midrule
\textbf{Frozen vs.\ executable} & Evidence classes are computed once and released
as a \emph{frozen, versioned} artifact---users never re-run the verifier, so this
layer is reproducible indefinitely. The judge must remain \emph{executable} to
score future submissions and is released as an open-weight model pinned by
revision, with its prompt, decoding parameters, and hashes. \\
\bottomrule
\end{tabular}
\end{table*}

\end{document}